\documentclass[10pt,twocolumn,letterpaper]{article}

\usepackage[pagenumbers]{wacv}%

\usepackage{pifont}%
\usepackage{multirow}%

\newcommand{\cmark}{\ding{51}}%
\newcommand{\xmark}{\ding{55}}%
\newcommand{\artgraph}{\ensuremath{\mathcal{A}rt\mathcal{G}raph}}

\usepackage[normalem]{ulem}
\newif\ifshowchanges
\showchangesfalse%
\ifshowchanges
  
  \newcommand{\del}[1]{\textcolor{revred}{\sout{#1}}}
  \newcommand{\moved}[1]{{\color{revgrey}\scriptsize[\,moved: #1\,]}}
\else
  
  \newcommand{\del}[1]{}
  \newcommand{\moved}[1]{}
\fi
\definecolor{revblue}{HTML}{1A4FBF}
\definecolor{revred}{HTML}{C42626}
\definecolor{revgrey}{HTML}{707070}
\ifshowchanges
  \newenvironment{newtext}{\color{revblue}}{}
\else
  
\fi

\definecolor{gaingreen}{HTML}{0F7B3C}
\definecolor{tiegrey}{HTML}{7A7A7A}
\newcommand{\gain}[1]{\textbf{\textcolor{gaingreen}{#1}}}
\newcommand{\tie}[1]{\textcolor{tiegrey}{#1}}

\newcommand{\snd}[1]{\underline{#1}}
\newcommand{\bestsnd}{Best in \textbf{bold}, second best \underline{underlined}.}

\usepackage[accsupp]{axessibility}%

\usepackage{qualfig}

\definecolor{wacvblue}{rgb}{0.21,0.49,0.74}
\usepackage[breaklinks,colorlinks,allcolors=wacvblue]{hyperref}

\def\wacvPaperID{303}
\def\confName{WACV}
\def\confYear{2027}

\title{ArtSeek: Deep artwork understanding via multimodal in-context reasoning and late interaction retrieval}

\author{Nicola Fanelli\,\orcidlink{0009-0007-6602-7504} \qquad Gennaro Vessio\,\orcidlink{0000-0002-0883-2691} \qquad Giovanna Castellano\,\orcidlink{0000-0002-6489-8628}\\
University of Bari Aldo Moro, Italy\\
{\tt\small \{nicola.fanelli, gennaro.vessio, giovanna.castellano\}@uniba.it}
}

\usepackage{orcidlink}

\usepackage[most]{tcolorbox}
\tcbuselibrary{listings,breakable}
\newtcblisting{promptbox}[1]{
    listing only,
    breakable,
    enhanced,
    colback=wacvblue!4,
    colframe=wacvblue,
    coltitle=white,
    fonttitle=\bfseries\small,
    title={#1},
    boxrule=0.6pt,
    arc=2pt,
    left=2mm, right=2mm, top=1mm, bottom=1mm,
    listing options={
        basicstyle=\ttfamily\scriptsize,
        breaklines=true,
        breakindent=0pt,
        columns=fullflexible,
        keepspaces=true,
    }
}

\newtcolorbox{prompttextbox}[1]{
    breakable,
    enhanced,
    colback=wacvblue!4,
    colframe=wacvblue,
    coltitle=white,
    fonttitle=\bfseries\small,
    title={#1},
    boxrule=0.6pt,
    arc=2pt,
    left=2mm, right=2mm, top=1mm, bottom=1mm,
    fontupper=\small,
}

\definecolor{wacvgreen}{rgb}{0.13,0.50,0.40}
\newtcolorbox{formattextbox}[1]{
    breakable,
    enhanced,
    halign=left,
    colback=wacvgreen!5,
    colframe=wacvgreen,
    coltitle=white,
    fonttitle=\bfseries\small,
    title={#1},
    boxrule=0.6pt,
    arc=2pt,
    left=2mm, right=2mm, top=1mm, bottom=1mm,
    fontupper=\small,
}
\newcommand{\ph}[1]{\texttt{\textless #1\textgreater}}

\begin{document}

\AddToShipoutPicture*{\AtPageLowerLeft{\put(\LenToUnit{0.75in},\LenToUnit{0.32in}){\parbox{\dimexpr\paperwidth-1.5in}{\centering\fontsize{6.5}{7.5}\selectfont Accepted at the IEEE/CVF Winter Conference on Applications of Computer Vision (WACV) 2027. \copyright\,2027 IEEE. Personal use of this material is permitted. Permission from IEEE must be obtained for all other uses, in any current or future media, including reprinting/republishing this material for advertising or promotional purposes, creating new collective works, for resale or redistribution to servers or lists, or reuse of any copyrighted component of this work in other works.}}}}
\maketitle
\begin{abstract}
Analyzing digitized artworks requires not only visual interpretation but also a deep understanding of rich artistic, contextual, and historical knowledge. We introduce ArtSeek, a multimodal framework for art analysis that combines multimodal large language models with retrieval-augmented generation. Unlike prior work, our pipeline relies only on image input, so it applies to artworks without links to Wikidata or Wikipedia---common in most digitized collections. ArtSeek integrates three components: a multimodal retrieval module based on late interaction retrieval, a contrastive multitask network predicting artist, genre, style, media, and tags, and an agentic reasoning strategy enabled through in-context examples. Central to this approach is WikiFragments, a dataset of image-text fragments extracted from the entire English Wikipedia, not just its art pages, to support knowledge-grounded multimodal reasoning. Our multimodal query encoding raises retrieval NDCG@5 from 33.9 to 44.9 over encoding the full image-text query, and fixed-budget token pooling keeps late-interaction search practical over millions of fragments at under 7\% of the cost of full multi-vector retrieval. ArtSeek sets a new state of the art in artwork attribute classification, with +8.4\% F1 in style classification over GraphCLIP, and gains +7.1 BLEU@1 on ArtPedia captioning. On five knowledge-grounded question answering benchmarks graded by an LLM judge from a different model family than the system under test, retrieval helps where the answer is documented in our index or lies inside the classifier's label space (+0.19 mean correctness on two of them) and is neutral where it is not; what separates the two regimes is how much of a benchmark our knowledge base covers. The gain transfers to two non-Qwen backbones (+0.22 to +0.28). A blind human study and a qualitative analysis of successes and failures corroborate this picture and expose where retrieval still misleads. Data and code: \href{https://cilab.di.uniba.it/artseek/}{\texttt{cilab.di.uniba.it/artseek}}.
\end{abstract}

\section{Introduction}
\label{introduction}

Analyzing and interpreting visual art remains one of the most compelling challenges in artificial intelligence. As Stork argues~\cite{stork2023pixels}, the difficulty is one of complexity rather than of principle: no inherent limitation stands between a computer and sophisticated art analysis. A system that could attribute a Renaissance painting, decode its symbolic elements, and situate it historically would have to integrate visual understanding, extensive knowledge, and complex reasoning---capabilities still beyond current systems. Resources such as our \artgraph{} knowledge graph~\cite{castellano2022leveraging} and methods built on it such as GraphCLIP~\cite{scaringi2025graphclip}, advance classification and retrieval but rely on fixed labels and limited context, while general-purpose models like CLIP~\cite{radford2021learning} and GPT-4V~\cite{achiam2023gpt} struggle with the art-historical knowledge and iconography that interpretation demands. MLLM-based efforts in the arts~\cite{yuan2023artgpt,bin2024gallerygpt} show potential but rely on parametric knowledge, risking shallow or hallucinated outputs.

We propose \textit{ArtSeek}, which unifies multimodal reasoning with retrieval-augmented generation (RAG) and, unlike systems such as KALE~\cite{jiang2024kale}, requires only an image as input---no metadata or links to external knowledge bases. This makes it applicable to the many artworks that lack annotations: of the 116{,}475 artworks in \artgraph{}~\cite{castellano2022leveraging}, only 13{,}626 link to Wikidata, and fewer still to Wikipedia. ArtSeek emulates the art historian's process through three core contributions:
\begin{itemize}
    \item \textit{Multimodal knowledge retrieval}: we curate \textit{WikiFragments}, a Wikipedia-scale dataset of image-text fragments, and apply late-interaction retrieval~\cite{khattab2020colbert} to fetch information about artists, symbols, and historical contexts.
    \item \textit{Attribute-aware classification}: our Late Interaction Classification Network (LICN) uses contrastive learning to jointly predict artist, style, genre, media, and tags, aligning visual features with label texts.
    \item \textit{Agentic reasoning}: with Qwen2.5-VL~\cite{bai2025qwen2} and an in-context example, the model plans retrieval, gathers supporting evidence, and synthesizes coherent interpretations.
\end{itemize}

By integrating classification, retrieval, and generation, ArtSeek improves on prior methods---e.g., +8.4\% F1 in style prediction over GraphCLIP~\cite{scaringi2025graphclip} and +7.1 BLEU@1 on ArtPedia~\cite{stefanini2019artpedia}---while producing richer, contextually grounded explanations. Beyond these, we contribute a knowledge-grounded evaluation of the full pipeline---five question answering benchmarks graded by an independent judge, three generator backbones, and a blind human study---reporting where it helps and where it does not, and measuring the coverage and label-space limits that account for the difference. Its non-parametric retrieval improves interpretability, letting users inspect the evidence behind an answer. While we instantiate ArtSeek only for visual arts, WikiFragments is extracted from the entire English Wikipedia; whether the framework transfers to other knowledge-intensive domains we leave to future work. We will release WikiFragments and our code.

\section{Related work}
\label{related}

\paragraph{Computer vision for art analysis} Deep learning is now central to art analysis, supported by large digitized collections such as WikiArt~\cite{fiorucci2020machine}. CNNs have been widely applied to attribute recognition and to feature-based retrieval and clustering~\cite{cetinic2018fine,van2015toward}, an unsupervised direction well-suited to unlabeled digital art and appealing to humanists seeking new interpretative patterns~\cite{wasielewski2023computational}. Since visual cues alone are often insufficient, later work injects context: ContextNet~\cite{garcia2020contextnet} fuses visual features with metadata or knowledge graphs, the \artgraph{} knowledge graph~\cite{castellano2022leveraging} encodes art-historical entities, and GraphCLIP~\cite{scaringi2025graphclip} aligns visual and contextual data via contrastive learning and graph neural networks.

\paragraph{Multimodal art analysis before LLMs} Datasets such as SemArt~\cite{garcia2018read} and ArtPedia~\cite{stefanini2019artpedia} paired artworks with text, enabling captioning and visual question answering~\cite{garcia2020dataset,cetinic2021iconographic,bai2021explain}, with some methods adding knowledge graphs for grounding~\cite{jiang2024kale}. These largely CNN--RNN models lacked the fluency and contextual depth of large pretrained language models.

\paragraph{LLMs and MLLMs for art} Recent work leverages (M)LLMs for richer and more flexible analysis~\cite{bongini2022gpt,castellano2023exploring,bin2024gallerygpt,yuan2023artgpt}, including connecting MLLMs to diffusion models for prompt-driven artwork inpainting~\cite{fanelli2025dream}; such systems typically combine a language model, a vision encoder~\cite{dosovitskiy2021an}, and alignment adapters~\cite{liu2023visual,achiam2023gpt}. To mitigate hallucinations, multimodal retrieval-augmented generation retrieves external knowledge~\cite{hu2023reveal,yan-xie-2024-echosight,caffagni2024wiki}; recent variants encode documents as images for vision-based retrieval~\cite{yu2025visrag} or plan retrieval with an agent~\cite{li2025benchmarking}. Unlike these, we extend a single model (ColQwen2~\cite{faysse2024colpali}) for training-free multimodal retrieval and repurpose the tool-calling ability of Qwen2.5-VL~\cite{bai2025qwen2} via in-context learning---inspired by ReAct~\cite{yao2023react}---to form multimodal sub-queries. An extended discussion is in \cref{related-full}.

\section{Data collection and the WikiFragments dataset}
\label{sec:dataset}

Answering art queries often requires external knowledge beyond the image itself, so a key component of ArtSeek is multimodal retrieval. State-of-the-art text retrieval encodes documents with multi-vector representations and \textit{late interaction}: ColBERT~\cite{khattab2020colbert} scores a query-document pair $(q,d)$ by matching each query token to its most similar document token (the \textit{MaxSim} operator) and summing,
\begin{equation}
S_{q,d}=\sum_{i\in[|E_q|]}\max_{j\in[|E_d|]}E_{q_i} \cdot E_{d_j},
\end{equation}
where $E_q$ and $E_d$ are the token-embedding sequences of the query and document. This is expressive but quadratic in cost and storage-heavy. ColPali~\cite{faysse2024colpali} extends it to multimodal retrieval by encoding document pages---including figures---as images with a VLM, outperforming multi-stage OCR/layout pipelines.

To retrieve knowledge for an input image and query, we build \textit{WikiFragments}. Rather than indexing whole pages as in ColPali, we treat individual passages as retrievable units: we adapt WikiExtractor~\cite{attardi2012wikiextractor} to extract from Wikipedia dumps both paragraphs and the images on a page (infobox and inline thumbnails with captions). A \emph{multimodal fragment} is a paragraph together with all images appearing above it in the page source. Each fragment is rendered as a single image---a four-cell-per-row grid of image-caption pairs (if any) followed by the paragraph text with hyperlinks preserved---so it can be encoded by ColQwen2 (a ColPali variant), whose embeddings jointly capture textual and visual content. Examples are shown in \cref{fig:fragment_examples} of the supplementary material.

Our pipeline yields over 42 million fragments covering English Wikipedia. As multi-vector representations are costly at this scale, we provide an algorithm that recursively traverses Wikipedia's category hierarchy to extract domain-specific subsets, and focus on the 8{,}166{,}323 fragments of the \textit{Visual Arts} category. We release this selection algorithm to support WikiFragments in other domains.

\section{ArtSeek}
\label{artseek}

\begin{figure}[t]
    \centering
    \includegraphics[trim=18pt 18pt 18pt 18pt, clip, width=\linewidth]{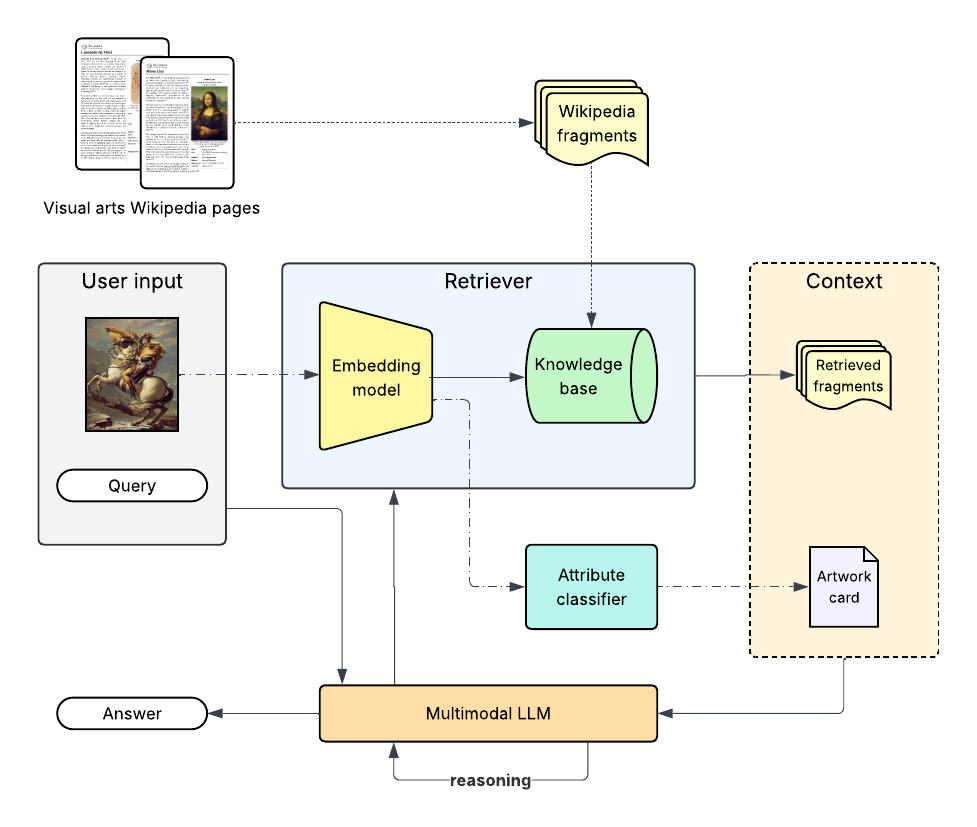}
    \caption{Overview of the ArtSeek pipeline. The retriever's embedding model doubles as a feature extractor for attribute classification (e.g., artist, style); the resulting artwork card is given to the MLLM, which learns via in-context learning to query the fragment knowledge base, enabling agentic RAG.}
    \label{fig:artseek}
\end{figure}

In this paper, we present \textit{ArtSeek}, a vision-language pipeline that integrates multimodal large language models with agentic retrieval-augmented generation to answer questions about artworks. As illustrated in Fig.~\ref{fig:artseek}, ArtSeek processes user queries about a given artwork image by addressing three core tasks: (1) multimodal retrieval for RAG, (2) artwork attribute classification, and (3) reasoning and text generation via an MLLM connected to the retriever. This reasoning capability is activated through in-context learning, using a single annotated example of a multi-turn conversation. The integration allows the model to identify when external knowledge is necessary and justify its answers and decision-making process. The following sections elaborate on our methodological contributions to each of these sub-tasks.

\subsection{Efficient multimodal retrieval with image-text inputs}
\label{sec:retrieval}

As shown in \cref{fig:retrieval} in the supplementary material, we leverage ColQwen2~\cite{faysse2024colpali}, a retrieval model built upon the Qwen2-VL-2B MLLM, to compute multi-vector representations of both queries and documents in our knowledge base. ColQwen2 is an adaptation of the original ColPali model, fine-tuned for retrieving image documents---potentially containing textual content---using text-only queries, enabling efficient and semantically rich late-interaction retrieval over multi-vector embeddings. However, a key limitation of ColQwen2 in our context is that it supports only text-based queries, whereas we aim for \textit{full} multimodal retrieval, where the user provides both an image (e.g., a digitized artwork) and a text query. Since many such queries reference specific visual elements of the image, purely textual retrieval---e.g., answering \textit{``Who is the man on the horse?''}---is likely to fail without incorporating visual information into the query.

\paragraph{Multimodal query encoding} To address this issue, we devise a strategy that enables ColQwen2 to process multimodal user queries by leveraging its internal token-level embedding structure. Given an image input $I = I_1 I_2 \ldots I_m$ and a text query $q = q_1 q_2 \ldots q_n$, we observe that their embeddings---used as input to the ColQwen2 language model---can be decomposed into meaningful subsequences:
\begin{equation}
I = I^{\text{pref}} \oplus I^{\text{content}} \oplus I^{\text{text}} \oplus I^{\text{suff}}, \quad
q = q^{\text{pref}} \oplus q^{\text{content}} \oplus q^{\text{suff}},
\end{equation}
where $\oplus$ denotes sequence concatenation. Each subsequence serves a distinct functional role in the model input, summarized in \cref{tab:embedding_structure} of the supplementary material.

Based on this structure, we construct a multimodal query representation by replacing the training-time prompt $I^{\text{text}}$ with the actual query $q$. This leads to the following composite sequence:
\begin{equation}
    q_{\text{multi}} = I^{\text{pref}} \oplus I^{\text{content}} \oplus q \oplus I^{\text{suff}}
\end{equation}
This formulation allows us to preserve the expected image-token structure while dynamically inserting user queries in place of the default text. The resulting multimodal sequence $q_{\text{multi}}$ is then passed to ColQwen2, which processes it through its transformer layers to produce a joint multimodal embedding:
\begin{equation}
E_{q_{\text{multi}}} = \mathrm{ColQwen2}(q_{\text{multi}})
\end{equation}
This design enables the model to jointly attend to visual and textual information, allowing more expressive and targeted multimodal reasoning.

However, we observed that this approach produces an excessively large number of embeddings for the query ($|E_{q_{\text{multi}}}| \gg |E_q|$), since images typically generate many input tokens. To address this, we retain only the embeddings corresponding to the textual part of the query from $E_{q_{\text{multi}}}$ after ColQwen2 processes the multimodal input. We refer to this subsequence as the \emph{filtered query embeddings}, denoted by $E_{q_{\text{multi}}}^{\text{filtered}}$. This strategy allows the model to leverage attention to enrich the textual embeddings with contextual visual information, effectively capturing image content relevant to the query. By using only these refined text-token embeddings, we substantially reduce the number of vectors involved in retrieval, which is crucial for the efficiency of multi-vector systems. Furthermore, as shown in Sec.~\ref{sec:retrieveres}, this selective approach significantly improves retrieval accuracy compared to using all embeddings.

\paragraph{Efficient document representation with token pooling} To further enhance efficiency, we optimize the way document embeddings are stored and processed. Given that our knowledge base $\mathcal{D}$ contains over 8 million multimodal fragments, building a fast retrieval index using full multi-vector representations becomes computationally prohibitive. To mitigate this, we adopt a \textit{token pooling strategy} inspired by recent work~\cite{clavie2024reducing}, aiming to reduce the number of embeddings per document without substantially affecting the quality of the retrieval phase.

Each document $d$, representing a multimodal fragment, is encoded as an image into a multi-vector representation $E_d$ using ColQwen2. This embedding sequence is partitioned into three parts: special token embeddings from the prefix ($E_d^{\text{pref}}$), content embeddings ($E_d^{\text{content}}$), and special token embeddings from the suffix ($E_d^{\text{suff}}$). We first compute the centroid of the special token embeddings, denoted as $c(E_d^{\text{pref}} \cup E_d^{\text{suff}})$. Then, we perform hierarchical clustering on the content embeddings $E_d^{\text{content}}$, grouping them into $K$ clusters, where $K$ is a small fixed budget (we set $K=8$). For each cluster, we compute its centroid to obtain a set of $K$ representative content vectors, denoted as $C_d$. Finally, we concatenate the special token centroid with the $K$ cluster centroids to construct the pooled representation of the document:
\begin{equation}
    E_d^{\text{pool}} = c(E_d^{\text{pref}} \cup E_d^{\text{suff}}) \oplus C_d
\end{equation}
where $\oplus$ denotes vector concatenation.

This results in a fixed-size representation of exactly $K + 1$ embeddings per document---one for the special tokens and $K$ for the clustered content---which we use for late interaction retrieval. Unlike prior work on token pooling for multi-vector search~\cite{clavie2024reducing}, which typically reduces the number of vectors by a constant factor (e.g., 2$\times$ or 4$\times$), our approach enforces a fixed embedding budget per document. This aggressive reduction makes the method highly scalable for large corpora, while experiments demonstrate that it still yields good retrieval performance, underscoring its practical effectiveness. To further enhance retrieval speed, we apply binarization to the pooled document embeddings, allowing us to store all representations in central memory for fast access. These binarized pooled embeddings serve as the first stage of our retrieval pipeline.

\paragraph{Two-stage retrieval process} Our retrieval pipeline consists of two stages:
\begin{enumerate}
    \item Initial retrieval: Using binarized pooled document embeddings, we efficiently identify the top $N_1$ candidate documents based on late interaction similarity.
    \item Re-ranking with full embeddings: We retrieve the full binarized multi-vector representations of the top $N_1$ documents (stored separately) and re-rank them based on fine-grained similarity matching with the query tokens.
\end{enumerate}
This two-stage approach strikes a balance between efficiency and precision, ensuring that the final retrieved documents are highly relevant to the multimodal user query. At the end of the process, we return $N_2$ documents, which our system deems most helpful in answering the query.

\subsection{Late Interaction Classification Network for multitask artwork card creation}

While MLLMs have a strong grasp of vision and language, their classification performance remains limited when labeled data is available. Although zero-shot or few-shot prompting can be used, training a dedicated classifier often yields better results when sufficient data exists. We find that MLLMs perform inconsistently when classifying attributes such as artist, genre, or style---especially for unfamiliar artworks. Since our framework must handle inputs absent from pretraining, we propose using metadata from \artgraph{}~\cite{castellano2022leveraging} to train a classifier that generates a metadata card for each artwork. This metadata card can then be passed to the MLLM, enriching its understanding and enabling more accurate, context-aware responses.

Specifically, we aim to classify five key attributes of an artwork: the \textit{artist}, \textit{genre}, and \textit{style}, which are multiclass classification tasks, as well as the \textit{media} used and the associated \textit{tags}, which are multilabel classification tasks. To tackle these classification tasks, we propose the Late Interaction Classification Network (LICN), a multitask learning architecture that uses ColQwen2 as its feature extractor. This addition integrates seamlessly into the existing retriever and MLLM pipeline within the RAG framework, enhancing classification capabilities with minimal system changes. An overview of LICN and its training process is shown in \cref{fig:licn} in the supplementary material.

Given an image $I$, our objective is to perform either \textit{multiclass classification}---assigning a single class---or \textit{multilabel classification}---assigning multiple relevant labels. We begin by encoding the image with ColQwen2, obtaining a multi-vector representation $E_I \in \mathbb{R}^{m \times 128}$, where $m$ is the number of image tokens and 128 is the dimensionality of each token embedding. To inject task-specific information, we prepend five learnable embeddings (one for each classification task) to $E_I$, resulting in an augmented sequence $F_I \in \mathbb{R}^{(5+m) \times 128}$. A transformer encoder then processes this sequence. From the output, we extract the embeddings corresponding to the five prepended task tokens and project them through a shared linear layer, yielding task-specific image embeddings $T_I \in \mathbb{R}^{5 \times 512}$. Each of these five embeddings encodes a 512-dimensional representation of the image tailored to a specific classification task.

In parallel, for each task $t\in \{1, \dots, 5\}$, we encode all candidate classes or labels $q$ (represented as text) using ColQwen2, obtaining their corresponding multi-vector representations $E_q$. We prepend a single, learnable task-specific token for each class or label (here, only the task embedding corresponding to the task to which the label pertains is prepended) and feed the resulting sequence into the same transformer encoder used for the image features. We then extract the transformed task token embedding and project it using the same linear layer, producing a consistent 512-dimensional representation $T_q \in \mathbb{R}^{512}$ for each class or label of the classification tasks.

To perform classification for task $t$, we compute the dot product between the task-specific image embedding and the embedding of each candidate class or label: $T_{I_t}^\top T_q$. We select the class $q$ with the highest score for multiclass classification. We apply a sigmoid function to the dot product for multilabel classification and compare it against a threshold to determine whether the label should be assigned to the image given as query.

We train the transformer encoder, the task embeddings, and the linear projection layer with a contrastive objective. For each task, the dot products between a batch's image embeddings and its candidate label embeddings form a logits matrix on which we apply a SigLIP-style sigmoid loss~\cite{zhai2023sigmoid} with learnable temperature and bias. The five per-task losses are combined through uncertainty weighting~\cite{kendall2018multi}, letting the model learn relative task weights; such multitask learning has been shown to benefit jointly predicting highly related artistic attributes such as artist and style~\cite{castellano2023exploring}. We give the full training formalism in \cref{impl-classification}.

\subsection{In-context learning for reasoning, retrieval, and answering with MLLMs}
\label{sec:generation}

As shown in Fig.~\ref{fig:artseek}, our pipeline centers on an MLLM, specifically Qwen2.5-VL-32B, chosen for its strong performance, its support for multiple image inputs, and its \textit{tool use}---the ability to invoke external functions during generation. Few open models at this scale pair vision-language quality with native tool use, and it is that pairing, rather than benchmark accuracy alone, that the agentic loop requires.

\paragraph{Motivation} Answering questions about artworks often requires different types of knowledge. Art historians distinguish between \textit{content}, \textit{form}, and \textit{context}~\cite{beltonart}, a taxonomy also adopted in~\cite{bai2021explain}. We find that Qwen2.5-VL performs well on visual aspects, but struggles with \textit{contextual queries} that require external knowledge, such as artist biographies or historical context. Additionally, queries like ``\textit{Describe the event depicted in this painting}'' are often too vague for direct retrieval. To address this, we adopt an \textit{agentic strategy} that allows the model to reason through the query, plan appropriate actions, and invoke external tools when needed.

\paragraph{Agentic strategy} We provide the retriever as a callable tool within the MLLM's reasoning loop. The process unfolds as follows:
\begin{itemize}
    \item Thinking: The model decomposes the query within special tokens \texttt{<think>} and \texttt{</think>}, following DeepSeek-R1~\cite{guo2025deepseek}. Generation starts after the \texttt{<think>} token.
    \item Tool use: The model formulates a retrieval query and decides whether to include the image:
    \begin{itemize}
        \item If needed, we use our retrieval method (Sec.~\ref{sec:retrieval}).
        \item Otherwise, we use ColQwen2~\cite{faysse2024colpali} for text-only retrieval.
    \end{itemize}
    \item Iteration: The model may repeat this process until a final answer is generated without any tool call.
\end{itemize}

\paragraph{Model steering via in-context examples} Attempts to steer the model via system prompts proved ineffective, likely due to Qwen2.5-VL's training on image-bound tool use. However, we found that the 32B model could generalize our strategy when shown a single, well-annotated \textit{in-context example}.

This example includes:
\begin{itemize}
    \item An image and user queries,
    \item The model's reasoning and tool calls,
    \item Tool outputs passed as messages, which the MLLM can ingest.
\end{itemize}

As illustrated in \cref{fig:generation} in the supplementary material, the example encompasses both simple visual questions and complex ones that require multiple retrieval steps. This allows the model to learn the full agentic reasoning loop and replicate it on new inputs.

In summary, ArtSeek differs from generalist large multimodal models such as ChatGPT by adapting a small, open MLLM with three lightweight mechanisms: an inherently multimodal late-interaction retriever, an agentic retrieval loop learned from a single annotated example without fine-tuning, and the LICN, which enriches the model's fine-grained classification---MLLMs are often unreliable at identifying the artist or style of a painting---by exploiting labeled metadata whenever it is available. The result rivals much larger systems while remaining simpler and more interpretable, as every statement is grounded in retrieved fragments the user can inspect.

\begin{table*}[t]
\centering
\scriptsize
\setlength{\tabcolsep}{5pt}
\begin{tabular}{l|ccc|ccc|ccc|c|c}
\hline
\textbf{Model} & \multicolumn{3}{c|}{\textbf{Artist}} & \multicolumn{3}{c|}{\textbf{Genre}} & \multicolumn{3}{c|}{\textbf{Style}} & \textbf{Media} & \textbf{Tags} \\
& Top-1 & Top-2 & F1 & Top-1 & Top-2 & F1 & Top-1 & Top-2 & F1 & F1 & F1 \\
\hline
ResNet \cite{castellano2022leveraging} & --- & --- & --- & 62.51 & 81.12 & 57.37 & 48.46 & 66.72 & 47.23 & --- & --- \\
ContextNet \cite{castellano2022leveraging, garcia2020contextnet} & --- & --- & --- & 61.77 & 79.87 & 56.70 & 42.61 & 61.91 & 41.42 & --- & --- \\
ViT+GAT \cite{castellano2022leveraging} & --- & --- & --- & 72.29 & 86.45 & 65.29 & 58.58 & 76.13 & 56.58 & --- & --- \\
ViT-Multitask \cite{castellano2023exploring} & \snd{69.93} & --- & \snd{58.63} & \snd{72.78} & --- & \snd{65.94} & 59.98 & --- & 57.41 & \snd{53.55} & \textbf{39.61} \\
GraphCLIP \cite{scaringi2025graphclip} & --- & --- & --- & 72.44 & \snd{86.62} & 65.06 & \snd{61.17} & \snd{77.81} & \snd{58.22} & --- & --- \\
LICN (Ours) & \textbf{71.75} & \textbf{80.59} & \textbf{65.01} & \textbf{78.54} & \textbf{89.34} & \textbf{73.25} & \textbf{69.80} & \textbf{84.23} & \textbf{66.65} & \textbf{59.49} & \snd{39.59} \\
\hline
\end{tabular}
\caption{Multitask classification results: Top-1/Top-2 accuracy and F1 for the multiclass tasks, F1 only for the multilabel ones. Since we adopt competing methods' reported results, some metrics are unavailable (``---''). \bestsnd}
\label{tab:classification1}
\end{table*}

\begin{table}[t]
\centering
\scriptsize
\setlength{\tabcolsep}{4pt}
\begin{tabular}{ll|ccccc}
\hline
\textbf{Dataset} & \textbf{Model} & B@1 & B@4 & M & S & R \\ \hline
\multirow{5}{*}{ArtPedia}    & Wu~\cite{wu2022artwork}           & 24.7  & 3.06 & 6.58 & --- & 22.4 \\
                             & Jiang et al.~\cite{jiang2024kale} & 32.6  & \textbf{7.48} & 9.33 & 7.68 & \textbf{23.7} \\
                             & Qwen2.5-VL-32B                    & 36.57 & 3.93 & 9.46 & 6.81 & 21.22 \\
                             & ArtSeek (w/o class.)              & \snd{39.64} & \snd{6.16} & \textbf{10.52} & \snd{7.73} & \snd{23.64} \\
                             & ArtSeek                           & \textbf{39.7} & 5.26 & \snd{10.26} & \textbf{7.74} & 23.0 \\ \hline
\multirow{5}{*}{SemArt v2.0} & Bai et al.~\cite{bai2021explain}  & ---   & \textbf{9.10} & \textbf{11.4} & --- & \textbf{23.1} \\
                             & Jiang et al.~\cite{jiang2024kale} & \snd{27.7}  & \snd{8.57} & \snd{9.51} & \textbf{7.31} & \snd{21.9} \\
                             & Qwen2.5-VL-32B                    & 26.49 & 1.02 & 5.91 & 3.58 & 16.89 \\
                             & ArtSeek (w/o class.)              & 27.45 & 1.09 & 5.98 & 3.97 & 16.66 \\
                             & ArtSeek                           & \textbf{28.15} & 1.30 & 6.21 & \snd{4.12} & 17.28 \\ \hline
\end{tabular}
\caption{Image captioning results, zero-shot for ArtSeek; some competing metrics are unavailable (``---''). PaintingForm~\cite{bin2024gallerygpt} follows the same pattern (\cref{tab:paintingform}), and BLEU@2--3 are reported in \cref{tab:bleu23}. \bestsnd B@N = BLEU@N, M = METEOR, S = SPICE, R = ROUGE-L.}
\label{tab:generation}
\end{table}

\section{Experiments}
\label{experiments}

We first evaluated the two components ArtSeek builds on in isolation---multimodal retrieval (\cref{sec:retrieveres}) and attribute classification (\cref{sec:classexp})---and then the assembled pipeline in three complementary ways: against prior art-captioning work under standard \textit{n}-gram metrics (\cref{sec:genexp}), on knowledge-grounded question answering with the answers graded for correctness (\cref{sec:vqa}), and with human raters on open-ended description (\cref{sec:human}). All experiments ran on one NVIDIA A100 with 64 GB of VRAM, with generation served by vLLM.

\subsection{Retrieval}
\label{sec:retrieveres}

We evaluated the strategy of \cref{sec:retrieval} on 10{,}000 WikiFragments, half text-only and half multimodal. Filtering the multimodal query down to its text tokens was the most effective choice: NDCG@5 rose from 33.92 to 44.88 and R@1 from 26.90 to 38.41 over the full-query encoding, at nearly half the query time. Combined with token pooling, it retained 27.61 NDCG@5 at under 7\% of the cost of full multi-vector retrieval, against 2.66 for a faster but much weaker CLIP baseline~\cite{baldrati2023composed} (\cref{impl-retrieval}).

\subsection{Classification}
\label{sec:classexp}

\subsubsection{Experimental setting}

We used the 116{,}475 WikiArt artworks of \artgraph{}~\cite{castellano2022leveraging}, each linked to one artist, genre, and style (multiclass) and optionally to tags and media (multilabel), split 70/15/15 and stratified by style, with ColQwen2 providing image and label embeddings. Following~\cite{castellano2023exploring} we report Top-1/Top-2 accuracy and F1 for artist, genre, and style, and F1 for tags and media. Our embedding-based formulation classifies over the full label set, restricted here only for fair comparison. Architecture and hyperparameters are in \cref{impl-classification}.

\subsubsection{Results and discussion}

Table~\ref{tab:classification1} compares LICN with prior work, including our own~\cite{castellano2022leveraging, castellano2023exploring, garcia2020contextnet, scaringi2025graphclip}. LICN outperforms all baselines, particularly on genre and style. We attribute this to using ColQwen2---a retriever pretrained on image-text data---as the feature extractor rather than ImageNet encoders fine-tuned on artworks, and to a contrastive setup whose batch-dependent loss improves generalization.
Operating in embedding space rather than with a fixed output layer, it also handles an open class set: it classifies any \artgraph{} artist (unlike~\cite{castellano2023exploring}) and adapts to new classes without architectural changes. An ablation (\cref{tab:classification2}) confirms that sharing the encoder and projection while keeping separate task embeddings works best. Per-head reliability is uneven and matters later: on held-out data the \textit{genre} and \textit{media} heads reach 80.6\% and 82.7\% top-1, while \textit{artist} reaches 29.6\% over 2501 classes.

\subsection{Artwork explanation}
\label{sec:genexp}

\subsubsection{Experimental setting}

We ran the pipeline with Qwen2.5-VL-32B (4-bit AWQ) as the MLLM, ColQwen2 multi-stage retrieval over all 8{,}166{,}323 WikiFragments ($N_1 = 100$, $N_2 = 10$) and LICN classification over all \artgraph{} labels. We evaluated zero-shot on ArtPedia~\cite{stefanini2019artpedia}, SemArt v2.0~\cite{bai2021explain}, and PaintingForm~\cite{bin2024gallerygpt} with standard captioning metrics, limited as they are for art explanation~\cite{bai2021explain}. Prompting and normalization details are in \cref{impl-generation} and \cref{prompts}.

\subsubsection{Results and discussion}

Table~\ref{tab:generation} compares against prior baselines~\cite{bai2021explain,jiang2024kale,wu2022artwork} and ablates the base MLLM, the retrieval-augmented variant, and the full pipeline. On ArtPedia, ArtSeek attains the best BLEU@1 and SPICE, and its retrieval-only variant the best METEOR. On SemArt v2.0 it leads in BLEU@1 but trails models trained on the dataset; high BLEU@1 with weaker higher-order \textit{n}-grams indicates correct vocabulary but different phrasing, reflecting a language bias~\cite{bleidt2024artquest}. 

The highest-scoring competitors are fine-tuned to reproduce each dataset's terse reference style, which \textit{n}-gram metrics reward while penalizing longer text. The improvement is nonetheless uneven, and \textit{n}-gram overlap cannot explain why: the subsections that follow set surface overlap aside and measure whether the answers are \emph{right}. 

\begin{table*}[t]
\centering
\scriptsize
\setlength{\tabcolsep}{5pt}
\begin{tabular}{ll|rc|ccc|c}
\hline
\textbf{Benchmark} & \textbf{Source} & \textbf{n} & \textbf{KB cov.} & \textbf{Qwen2.5-VL-32B} & \textbf{ArtSeek (w/o class.)} & \textbf{ArtSeek} & \textbf{ArtSeek $-$ backbone [95\% CI]} \\
\hline
\multicolumn{8}{l}{\textit{Paintings the knowledge base covers}} \\
ArtPedia-VQA  & ours                                    & 651  & \gain{85.8\%} & 0.535 & \textbf{0.774} & \snd{0.722} & \gain{+0.186} [+0.119, +0.254] \\
LICNHeldOut   & ours (Wikidata)                         & 762  & \gain{53.3\%} & 1.010 & \snd{1.086} & \textbf{1.197} & \gain{+0.186} [+0.114, +0.257] \\
\hline
\multicolumn{8}{l}{\textit{Paintings the knowledge base barely covers}} \\
AQUA          & Garcia et al.~\cite{garcia2020dataset}  & 652  & \tie{10.4\%} & \textbf{0.920} & \snd{0.910} & 0.896 & \tie{$-$0.025 [$-$0.095, +0.046]} \\
ArtCurate-AIC & ours (AIC, CC0)                         & 1489 & \tie{10.3\%} & \textbf{1.125} & 1.074 & \snd{1.093} & \tie{$-$0.034 [$-$0.071, +0.003]} \\
ArtQuest      & Bleidt et al.~\cite{bleidt2024artquest} & 1200 & \tie{\phantom{0}8.9\%} & \textbf{0.937} & 0.828 & \snd{0.914} & \tie{$-$0.023 [$-$0.076, +0.031]} \\
\hline
\end{tabular}
\caption{Question answering: mean correctness on a 0--2 rubric. ``KB cov.'' is the share of a benchmark's paintings having any page in our index; \gain{green} marks a difference whose interval excludes zero, \tie{grey} one that spans it; \textbf{bold} marks the best system in a row and \underline{underlining} the second best. The table splits by coverage rather than by benchmark family: ArtSeek gains where the knowledge base covers the painting, and is indistinguishable from the bare backbone where it does not.}
\label{tab:vqa}
\end{table*}

\begin{figure}[t]
    \centering
    \input{figures/qual/qual_r2}
    \caption{Why the question alone is not a usable retrieval query: it names no painting, no artist, no period and no place, so embedding it returns generic articles about headwear. The image and the artwork card supply what the query needs, and the backbone's confident ``hennin'' becomes the correct \textit{balzo}---even though the card's artist field is wrong here. Correct identifications are green, incorrect ones red.}
    \label{fig:qual_r2}
\end{figure}

\subsection{Knowledge-grounded question answering}
\label{sec:vqa}

\paragraph{Setting} Captioning metrics measure surface overlap, not truth, so we assembled five question answering benchmarks and graded correctness directly (\cref{tab:vqa}; construction in \cref{r2-benchmarks}). ArtPedia-VQA pairs each ArtPedia test painting with one visual and one contextual question from its human-authored sentences. AQUA~\cite{garcia2020dataset} and ArtQuest~\cite{bleidt2024artquest} are external and published, the latter built to remove language bias. LICNHeldOut asks Wikidata-derived questions about paintings by LICN's 2501 artist classes, excluding every \artgraph{} training artwork; no language model wrote its references. ArtCurate-AIC asks conceptual questions---iconography, narrative, technique, context---from CC0 museum catalogue prose rather than factual lookups.

Each answer was graded against the benchmark's reference answer on a three-point correctness rubric: \textbf{2} when it conveys the reference's key facts, however it is phrased and however much it elaborates; \textbf{1} when it gets some of them right but misses, hedges or mixes in wrong ones; and \textbf{0} when it contradicts them or declines to answer---a refusal scores 0 because the reference itself shows the answer was determinable. Mean correctness on this scale is what we report throughout. The grader is phi-4~\cite{abdin2024phi4}, from a model family different from the generator's and from those used to build the benchmarks, so no part of the evaluation loop shares a lineage with the system it grades. Every comparison is paired over questions and bootstrapped over paintings (10{,}000 cluster resamples), so a painting with several questions counts once; we report 95\% confidence intervals throughout.

\paragraph{Coverage decides the outcome} \Cref{tab:vqa} splits in two, but not by provenance or question style. What separates the benchmarks where ArtSeek gains from those where it does not is how much of each our index covers. Where most paintings have a page, retrieval was worth about $+0.19$ mean correctness; where almost none do, ArtSeek and the bare backbone cannot be told apart.

\paragraph{The artwork card has the scope of its label space} The card behaves like the classifier it is, helping where its label space contains the answer and hurting where it does not. LICNHeldOut was built to test exactly that: its questions come from Wikidata properties that map one-to-one onto LICN heads, on paintings the classifier never saw. There the card carries the larger part of the gain, adding $+0.111$ to the $+0.076$ retrieval wins on its own. ArtQuest shows the same from the other side, its question types matching LICN's heads only in part: the card was worth $+13.5$\,pp on \textit{technique} and $+7.0$\,pp on \textit{type}, and cost $-11.5$\,pp on \textit{school}, for which no head exists---a pattern we predicted from LICN's vocabularies before the run (\cref{r2-extra}).

\paragraph{The gain is not tied to one backbone} We re-ran the pipeline on Gemma-3-27B~\cite{gemmateam2025gemma3} and Mistral-Small-3.1-24B~\cite{mistral2025small31}, holding retriever, index, classifier and prompt fixed, so any change could only come from the generator. Retrieval helped on all three, by $+0.22$ to $+0.28$, and most on the weakest, so it is not merely amplifying an already strong model. Neither alternate can emit tool calls, so both ran with retrieval seeded before the first turn. That same limitation is why we build on Qwen2.5-VL rather than on whichever backbone scores highest alone (\cref{r2-serving}).

A gain of this size invites a memorization check. Shown a painting alone, Qwen2.5-VL-32B identified neither title nor artist for 180 of the 326 ArtPedia works; on those, retrieval still gained $+0.200$ [$+0.117$, $+0.286$], so the improvement survives on works the model does not know.

\paragraph{Does the in-context policy earn its place?} To test whether the policy learned from a single worked example does any real work, we replaced that example with a prose description of the same policy, changing nothing else. The demonstration won on all five benchmarks, by $+0.05$ to $+0.19$ mean correctness, and four of the five intervals exclude zero. The exception is AQUA, whose source corpus our index does not contain: with no evidence to retrieve, no policy has room to distinguish itself. It won on cost too: the prose variant was 52\% slower, issued more retrieval calls to reach a lower score, and sometimes emitted no final answer at all. The prose prompt is also the longer, so the example is not winning by saying more (\cref{tab:policydelta}).

Grading the retrieved fragments rather than the final answers shows where the main room for improvement lies. An oracle that re-queries the index using the reference answer reduces ArtPedia's empty-evidence rate from about one half to one tenth. This suggests that the main issue is recognition rather than query phrasing. For example, the question ``what is the figure on the left holding?'' does not mention anything specific that the retriever can match. If the system recognizes the object as a martyr's palm, it can instead retrieve the article about the saint who carries one---something that simply rephrasing the question would not achieve. This means that retrieval benefits directly from better visual understanding: every object, face, or costume the backbone can recognize provides a term that the retriever can use to find relevant documents (\cref{fig:qual_r2}). 

\paragraph{Cost} A full answer took 11.8\,s at the median against 6.5\,s for the backbone alone. Classification accounts for 0.14\,s of that, so the artwork card is a negligible share of the added cost: retrieval and generation account for the rest. These timings come from a \emph{shared} A100 and are upper bounds, not tuned deployment figures (\cref{tab:latency}).

\subsection{Human evaluation}
\label{sec:human}

Answering questions correctly is not the same as describing a painting well, and it is the description a gallery visitor actually reads. So we put that task to people. Each system wrote a four-part description of a painting---subject, figures, placement, and the evidence its reading rests on---scored on the same 0--2 scale, applied to what is \emph{claimed} rather than to how well it is written. Three raters, two lay and one an art historian, each scored 14 paintings across Qwen2.5-VL-32B, ArtSeek and a tool-free GPT-5.5, shown as A/B/C in an order differing per rater and per painting. All descriptions were first rewritten into one neutral register by the judge model, so raters compared claims, not prose style. Ten paintings have a Wikipedia article a frontier model may have memorized; the other four, from the Galleria Nazionale della Puglia in Bitonto, have none and cannot be reproduced here.

ArtSeek scored above its backbone on the aggregate ($\Sigma$-correctness 3.71 against 3.02 on a 0--6 scale), on both tiers and all but one axis, and came within 0.17 of the tool-free frontier model at 3.88. The one difference this study can certify falls where it should: on the four undocumented works ArtSeek gained $+1.17$ [$+0.50$, $+2.00$] over its backbone, which collapses when the painting has no article while ArtSeek barely moves. On those same works ArtSeek drew level with the frontier model ($+0.17$ [$-2.17$, $+2.50$], the forced choice splitting evenly), which we read as parity with a much larger closed system, not a win. The study also validates the automatic judge: phi-4 tracks the raters at least as closely as they track each other (\cref{r2-human}).

\subsection{Qualitative analysis}
\label{sec:qual}

\Cref{fig:qual_r2} makes the same point on a single item, with the card working as a query aid rather than an oracle: its artist field is wrong there while the answer is right. Two failure modes deserve naming. Retrieval can overturn an answer that was already correct, by surfacing a related work and carrying its facts over to the painting at hand; and when the corpus lacks the answer, the backbone invents one while ArtSeek declines. Both score zero under our rubric; a curator would not treat them alike. \Cref{qualitative} reports full traces for both.  

\section{Conclusion}
\label{conclusion}

This paper introduced ArtSeek, which combines agentic retrieval-augmented generation with attribute-aware classification. From visual input alone it couples WikiFragments---8.2 million art-related image-text fragments---with training-free late-interaction retrieval, the Late Interaction Classification Network for attribute prediction, and an in-context agentic loop that plans retrieval, so it can analyze any artwork while exposing its sources. It gains +8.4\% F1 in style classification over GraphCLIP and +7.1 BLEU@1 on ArtPedia, and improves knowledge-grounded question answering wherever the answer is documented in the index or lies inside the classifier's label space. Its limits are stated as carefully: retrieval is flat where coverage is low, and a refusal scores like a fabrication. Future work includes expanding the knowledge base beyond Wikipedia, exploiting WikiFragments hyperlinks for multi-hop navigation, learning the retrieval policy rather than demonstrating it, and building evaluation protocols with art historians.

\section*{Societal impact}
ArtSeek can make art-historical knowledge more accessible to museums, educators, and the public, including for the many artworks that lack catalogue metadata. It is built entirely on open-weight models and an explicit knowledge base, so it can be inspected and run locally, and institutions can extend or correct the sources it retrieves from rather than relying on a closed model's parametric knowledge. That knowledge base, however, inherits English Wikipedia's coverage, which favours Western and well-documented art; as our results show, retrieval helps even on works the backbone does not recognize, but only when they are documented in the index, so the system may serve underrepresented traditions worse. Generated attributions and explanations can also be fluent yet wrong, which matters where they inform scholarship, conservation, or valuation. ArtSeek exposes its retrieved sources so that claims can be verified, but its outputs must be checked by domain experts: they are meant to support expert judgement, not replace it.

\section*{Acknowledgements}
We acknowledge ISCRA for awarding this project access to the LEONARDO supercomputer, owned by the EuroHPC Joint Undertaking, hosted by CINECA (Italy). Nicola Fanelli's research is funded by a Ph.D. fellowship under the Italian ``D.M. n. 118/23'' (NRRP, Mission 4, Investment 4.1, CUP H91I23000690007). We thank the Galleria Nazionale della Puglia ``Girolamo e Rosaria Devanna'' (Bitonto, Italy) for the paintings used in our human study, and the member of its staff who took part as a rater.

{
    \small
    \bibliographystyle{ieeenat_fullname}
    \bibliography{references}
}

\clearpage
\setcounter{section}{0}\setcounter{figure}{0}\setcounter{table}{0}\setcounter{equation}{0}
\renewcommand{\thesection}{S\arabic{section}}
\renewcommand{\thefigure}{S\arabic{figure}}
\renewcommand{\thetable}{S\arabic{table}}
\renewcommand{\theequation}{S\arabic{equation}}
\maketitlesupplementary

\section{Extended related work}
\label{related-full}

This section provides the full discussion of related work, condensed in the main paper for space reasons.

\subsection{Deep learning and computer vision for art analysis}
\label{cvart}

The rise of large datasets and accessible computational power has solidified deep learning, particularly deep neural networks, as the leading approach for complex tasks in computer vision and NLP. This shift extends to cultural heritage, aided by institutions digitizing vast artwork collections and platforms like WikiArt~\cite{fiorucci2020machine} providing public access to visual data.

Convolutional neural networks (CNNs) have been widely used for attribute recognition in art images~\cite{cetinic2018fine,van2015toward}, achieving performance levels previously unattainable. Other works leverage CNN-extracted features with distance metrics for tasks like artwork retrieval and clustering. These efforts highlight the promise of unsupervised learning, which is well-suited for the abundance of unlabeled digital art and appealing to humanists interested in uncovering new interpretative patterns~\cite{wasielewski2023computational}.

Nevertheless, understanding visual art requires more than image data---it also depends on contextual information. Garcia et al.~proposed ContextNet~\cite{garcia2020contextnet}, which enriches CNN-derived visual features with contextual data through multi-task learning or knowledge graphs. Both strategies led to improved results in classification and retrieval tasks. The integration of knowledge graphs, in particular, has proven effective in bridging the gap between visual and semantic information. For instance, Castellano et al.~introduced \artgraph{}~\cite{castellano2022leveraging}, a domain-specific knowledge graph encoding entities such as artists, artistic styles, and historical periods. Building on this, Scaringi et al.~developed GraphCLIP~\cite{scaringi2025graphclip}, which aligns visual and contextual data for style and genre classification using contrastive learning and graph neural networks, offering interpretability via node importance.

\subsection{Multimodal art analysis before large language models}
\label{artvl-prellm}

While attribute recognition in artworks provides valuable insights, it lacks the richness offered by multimodal approaches that integrate visual and textual information. Early work introduced datasets that combine images with text to support semantic and cross-modal retrieval. SemArt~\cite{garcia2018read} was the first to include artistic commentary alongside painting attributes, followed by ArtPedia~\cite{stefanini2019artpedia}, which incorporated numerous artwork descriptions from both visual and contextual perspectives.

These datasets enabled tasks such as image captioning and visual question answering (VQA) in the artistic domain. Garcia et al.~\cite{garcia2020dataset} developed AQUA to distinguish between visual and knowledge-based questions, while Cetinic~\cite{cetinic2021iconographic} generated iconographic captions using Iconclass annotations. Bai et al.~\cite{bai2021explain} enriched SemArt with content, form, and context labels, combining DrQA, object detection, and attribute prediction. Other strategies, such as KALE~\cite{jiang2024kale}, integrated knowledge graphs to enhance factual grounding and disambiguation.

Most of these architectures followed an encoder-decoder paradigm, typically using a CNN for visual encoding and a recurrent neural network (RNN) for textual generation. Some later works adopted transformer-based models for either modality, but none leveraged large pretrained language models trained on massive corpora. As a result, their ability to generate contextually rich and fluent text was limited, particularly for abstract or culturally loaded visual inputs.

\begin{figure*}[!t]
    \centering
    \includegraphics[width=1.0\linewidth]{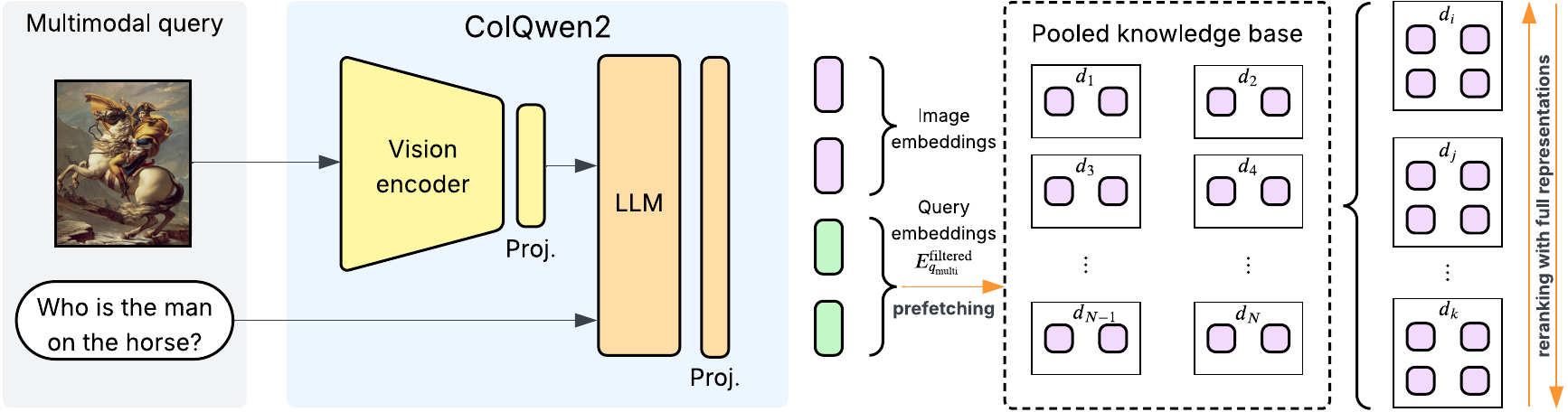}
    \caption{Our multi-stage retrieval pipeline. We design an efficient retrieval method over a large-scale knowledge base $\mathcal{D} = \{d_i\}_{i=1}^N$ using a multi-vector strategy and multimodal queries to ColQwen2, without compromising efficiency or worsening the modality gap. Both the input image and text are encoded with ColQwen2, but only text-related embeddings are used for late interaction retrieval. The first stage retrieves $N_1$ candidates via pooled representations; the second reranks them using full multi-vector representations to return the top $N_2$ results ($N_2 < N_1$).}
    \label{fig:retrieval}
\end{figure*}
\begin{figure*}[!t]
    \centering
    \includegraphics[width=1.0\linewidth]{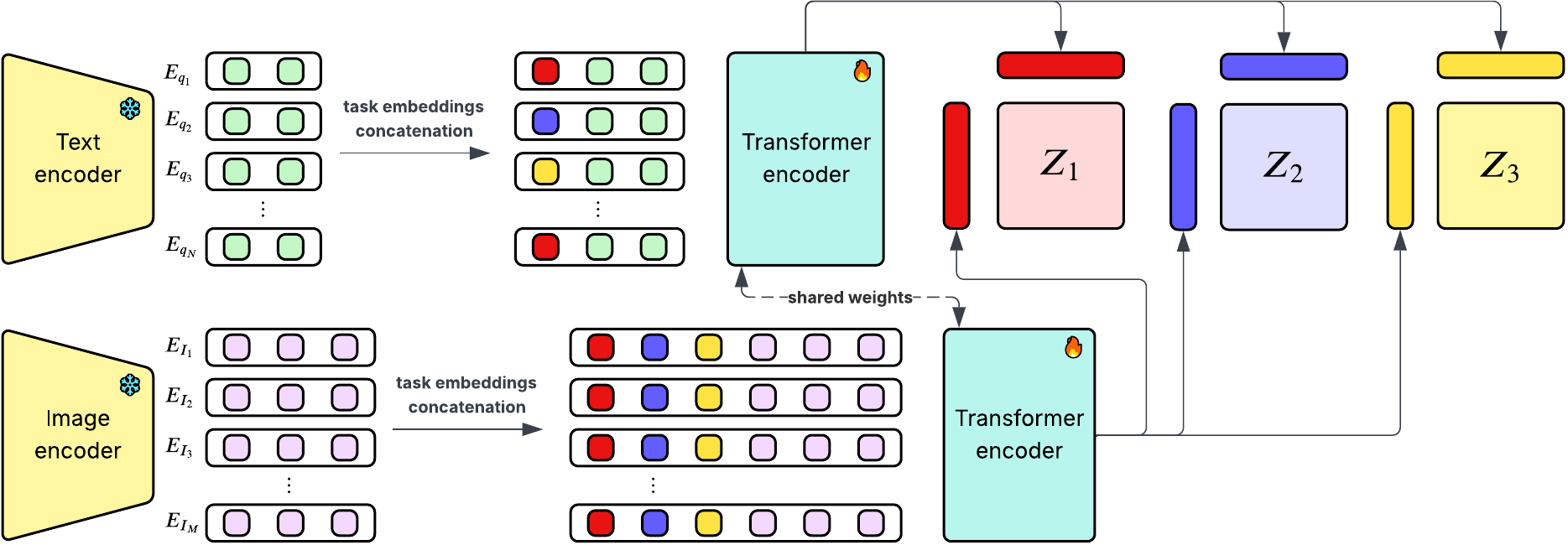}
    \caption{Overview of the contrastive training process for LICN. Given a batch of images $\{I_i\}_{i=1}^M$ with corresponding labels $\{q_i\}_{i=1}^N$ for multiple classification tasks, we prepend multiple task tokens to each image's multi-vector representation $E_{I_i}$ and a single task token to each label encoding $E_{q_i}$. Both are processed by a transformer, producing task-aware embeddings by extracting the contextualized task tokens. We then compute a sigmoid loss on the per-task logits matrix $Z_t$, whose entry $(i,j)$ is the dot product between image $i$'s and label $j$'s task-$t$ embeddings; there is one such matrix per task, of which three are drawn here.}
    \label{fig:licn}
\end{figure*}

\subsection{Multimodal art analysis with LLMs and MLLMs}
\label{artvl-llm}

The emergence of large language models (LLMs) and multimodal LLMs (MLLMs) has opened new opportunities for artwork analysis by improving the fluency, flexibility, and contextual awareness of generated content.

Bongini et al.~\cite{bongini2022gpt} explored the use of ChatGPT for cultural heritage question answering, noting both its effectiveness and its limitations due to hallucinations. Castellano et al.~\cite{castellano2023exploring} proposed generating synthetic artwork descriptions with ChatGPT, improving training with CLIP-based weighting. GalleryGPT~\cite{bin2024gallerygpt} extended this line by combining MLLMs with art-specific annotation tools like PaintingForm, showing the potential for zero-shot generalization in formal analysis. Similarly, ArtGPT4~\cite{yuan2023artgpt} fine-tuned MiniGPT4 on a large corpus of web-scraped art data. Beyond analysis, MLLMs have also been connected to diffusion models to drive generative editing of paintings, as in a recent prompt-generation approach~\cite{fanelli2025dream} for text-guided multi-mask inpainting.

Many of these approaches use standard MLLM architectures comprising a pre-trained language model, a vision encoder (e.g., ViT~\cite{dosovitskiy2021an}), and adapter modules for modality alignment. Notable models include LLaVA~\cite{liu2023visual} and GPT-4V~\cite{achiam2023gpt}, often trained using two-stage protocols and large-scale multimodal datasets like LAION-2B.

To overcome the limits of parametric memory, retrieval-augmented generation (RAG) has been extended to multimodal tasks. REVEAL~\cite{hu2023reveal} and EchoSight~\cite{yan-xie-2024-echosight} retrieve knowledge based on visual queries, combining them with reranking strategies. Wiki-LLaVA~\cite{caffagni2024wiki} incorporates hierarchical pipelines based on CLIP-based entity linking and dense passage retrieval. More recent systems push multimodal RAG further: VisRAG~\cite{yu2025visrag} encodes documents directly as images to avoid lossy parsing, while Li et al.~\cite{li2025benchmarking} introduce a self-adaptive planning agent that dynamically decides what and when to retrieve for complex knowledge-seeking VQA.

In contrast, our method encodes the full (image + question) input by extending a single model (ColQwen2~\cite{faysse2024colpali}) without any additional training. Specifically, we build on Qwen2.5-VL~\cite{bai2025qwen2}, which integrates a ViT-based vision encoder with a large-scale language model and supports \textit{tool calling}. We repurpose this mechanism to perform open-domain multimodal retrieval, allowing the model to use the retriever to query a knowledge base containing both text and image fragments.

Rather than relying on handcrafted retrieval pipelines, we design prompt templates that guide the model to decompose complex user queries into one or more multimodal sub-queries. This strategy is inspired by frameworks such as ReAct~\cite{yao2023react}, but implemented entirely via in-context learning. In doing so, we awaken the latent planning and reasoning capabilities of the model, enabling adaptive retrieval based on the user question.

\section{Method diagrams}
\label{supp:method}

This section collects the architectural diagrams referenced in the main paper (\cref{artseek}), together with the structure of the multimodal input embeddings that \cref{sec:retrieval} builds on. \Cref{fig:retrieval} illustrates our multi-stage multimodal retrieval pipeline (\cref{sec:retrieval}), and \cref{fig:licn} illustrates the contrastive training process of the Late Interaction Classification Network (LICN).

\subsection{Structure of the multimodal input embeddings}
\label{sec:embedding_structure}

\Cref{tab:embedding_structure} details the functional subsequences into which
ColQwen2's image and query inputs decompose, and which our multimodal query
encoding recombines (\cref{sec:retrieval}).

\subsection{WikiFragments examples}
\label{sec:fragment_examples}

\begin{figure}[t]
    \centering
    \begin{subfigure}[t]{\linewidth}
        \centering
        \fbox{\includegraphics[width=\linewidth]{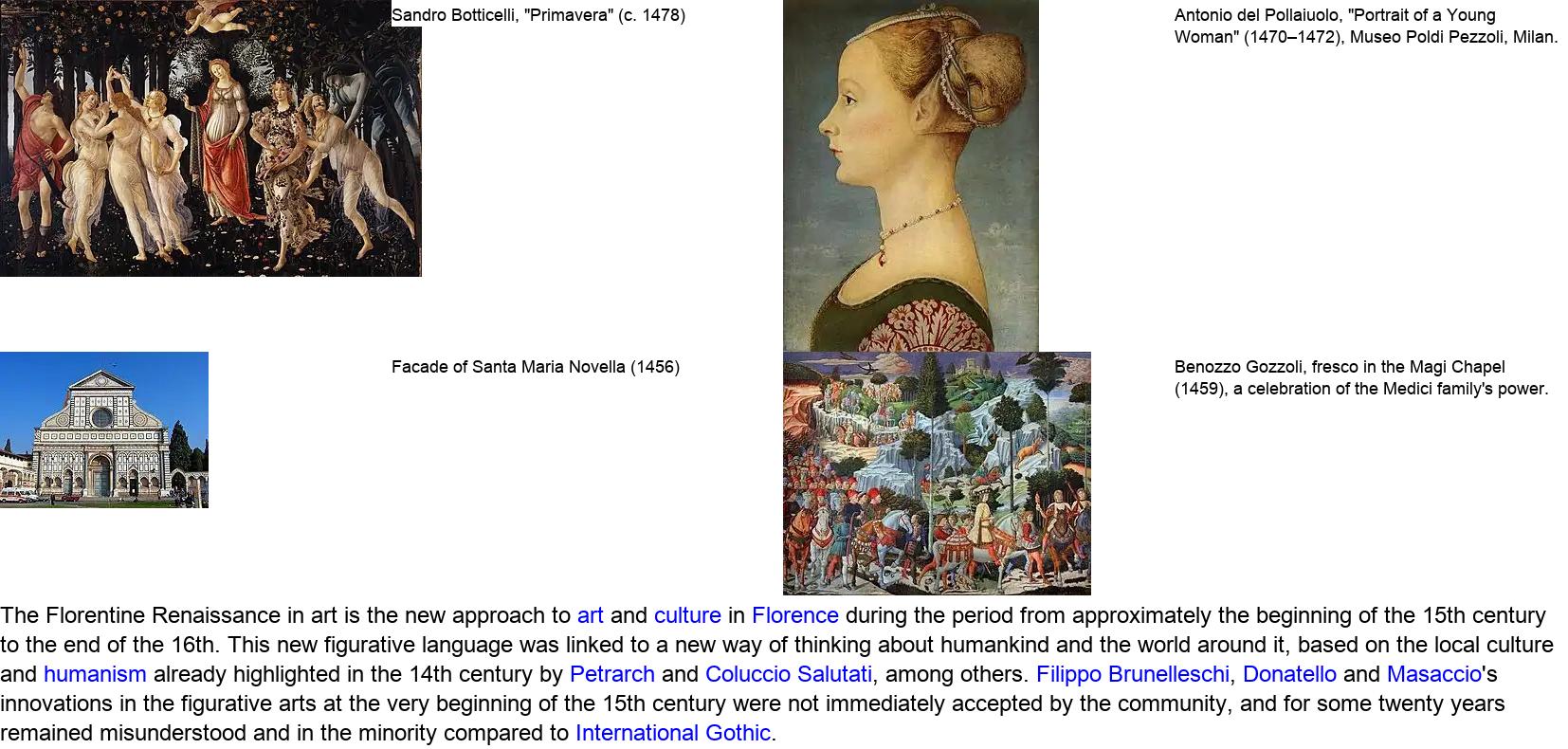}}
        \caption{Fragment with four images and captions}
    \end{subfigure}

    \vspace{0.6em}

    \begin{subfigure}[t]{\linewidth}
        \centering
        \fbox{\includegraphics[width=\linewidth]{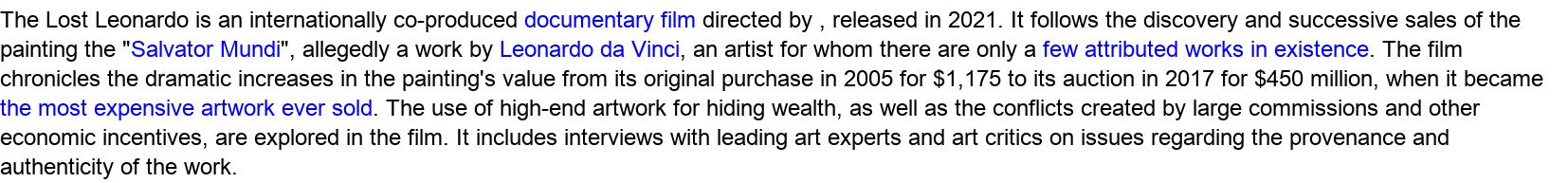}}
        \caption{Fragment with only text and no associated images}
    \end{subfigure}

    \caption{Examples from the WikiFragments dataset. \textbf{(a)} A fragment with four images, their captions, and the paragraph text. \textbf{(b)} A text-only fragment. For each fragment we also store metadata about the source Wikipedia page, its hyperlinks, and the paragraph's position within the page.}
    \label{fig:fragment_examples}
\end{figure}

\Cref{fig:fragment_examples} shows two entries of the WikiFragments dataset
(\cref{sec:dataset}): a fragment carrying four images with their captions, and a
text-only fragment.

\begin{table}[t]
\centering
\resizebox{\columnwidth}{!}{%
\begin{tabular}{ll}
\toprule
\textbf{Subsequence} & \textbf{Description} \\
\midrule
$I^{\text{pref}}$      & Embeddings for the preamble introducing the image input. \\
$I^{\text{content}}$   & Embeddings for the image patches, surrounded by \texttt{<|vision\_start|>} and \texttt{<|vision\_end|>}. \\
$I^{\text{text}}$      & Embeddings of the default prompt ``\textit{Describe the image.}'' used during training. \\
$I^{\text{suff}}$      & End-of-message special tokens for image input. \\
$q^{\text{pref}}$      & Embeddings for the prefix string ``\textit{Query: }''. \\
$q^{\text{content}}$   & Embeddings for the actual user query. \\
$q^{\text{suff}}$      & Special end-of-query tokens. \\
\bottomrule
\end{tabular}
}
\caption{Structure of multimodal input embeddings. Each input (image or text) is represented as a sequence of embeddings composed of functional subsequences.}
\label{tab:embedding_structure}
\end{table}

\section{Implementation and experimental details}
\label{impl}

This section reports the full implementation and experimental setup, summarized in the main paper for space reasons.

\subsection{Retrieval evaluation}
\label{impl-retrieval}

We selected 10{,}000 fragments from our WikiFragments dataset to evaluate retrieval performance. Specifically, we chose 5{,}000 fragments originally containing a single image and 5{,}000 with at least two images. In all cases, we removed the first image from each fragment. This resulted in 5{,}000 text-only fragments and 5{,}000 fragments that retained one or more images, allowing us to assess retrieval over both text-only and multimodal documents.

To build multimodal queries, we used the removed images and prompted Qwen2.5-VL-32B to generate a question about each image that could be answered using the corresponding paragraph. To ensure grounding in the visual content, all prompts required the generated question to include the phrase ``\textit{this image}''.

We created two data stores for retrieval. The first stores full multi-vector representations for all 10{,}000 fragments. The second applies our retrieval strategy: prefetching based on pooled token representations followed by reranking using full fragment embeddings (up to 768, as in the original ColPali paper). Both use binary quantization and set vector oversampling to $2$. Retrieval was conducted using Qdrant with default HNSW index settings. For the multi-stage retrieval setup, we set the number of clusters for token pooling to $K = 8$. In the first stage, we retrieve $N_1 = 100$ candidate passages, which are then reranked to select the top $N_2 = 5$ results. For full retrieval, the top 5 documents are returned directly in a single stage.

We evaluate performance using Normalized Discounted Cumulative Gain (NDCG) and Recall (R). NDCG rewards the early ranking of relevant results, while Recall measures whether the original fragment (from which the query was derived) was successfully retrieved. We also report execution time (in centiseconds) to compare retrieval efficiency.

\Cref{tab:retrieval} reports the full results, summarized in \cref{sec:retrieveres}
of the main paper. Using full multi-vector representations for both store and query
is accurate but slow---already costly at 10{,}000 fragments and impractical at the
scale of millions. The CLIP baseline, which sums query image and text
embeddings~\cite{baldrati2023composed}, is the fastest thanks to single-embedding
retrieval, but very weak: its speed comes at a steep cost in accuracy. This reflects
both CLIP's limited ability to retrieve fragments whose relevance is textual rather
than visual, and the advantage of building retrieval on an LLM backbone like
ColQwen2, whose far broader world and contextual knowledge lets it match queries to
fragments a CLIP-style encoder cannot. Our multimodal query encoding, which retains
only question-relevant tokens, achieves the best NDCG@5 and R@1 while nearly halving
query time, since fewer query tokens enter the quadratic late-interaction step. Token
pooling with \emph{full} queries performs worst, as document tokens are then far
outnumbered by query tokens, wasting MaxSim operations; but combining token pooling
with filtered queries matches full retrieval at under 7\% of its cost. We therefore
recommend full representations only for small corpora and the two-stage pooled
strategy for large-scale deployment.

\begin{table}[t]
\centering
\small{
\resizebox{\columnwidth}{!}{%
\begin{tabular}{ll|c|c|c}
\hline
\textbf{Store}   & \textbf{Query}   & \textbf{NDCG@5} (\textuparrow) & \textbf{R@1} (\textuparrow) & \textbf{Avg time (cs)} (\textdownarrow) \\ \hline
CLIP    & CLIP    & 2.66       & 1.49    & \textbf{0.39} \\
Full (one-stage)    & Full ($E_{q_{multi}}$)    & \snd{33.92}      & \snd{26.90}   & 67.92        \\
Full (one-stage)   & Filtered ($E_{q_{multi}}^{{\text{filtered}}}$) & \textbf{44.88}      & \textbf{38.41}   & 32.26        \\
Pooled (two-stage) & Full ($E_{q_{multi}}$)    & 21.50      & 17.39   & 11.14        \\
Pooled (two-stage) & Filtered ($E_{q_{multi}}^{\text{filtered}}$)  & 27.61      & 23.87   & \snd{4.66}        \\ \hline
\end{tabular}
}
}
\caption{Retrieval results, including NDCG@5 and Recall@1 (R@1), where ``@K'' indicates evaluation over the top $K$ retrieved items, and the average query time required for retrieval. \bestsnd}
\label{tab:retrieval}
\end{table}

\subsection{Classification training}
\label{impl-classification}

Classification was performed on 116{,}475 artwork images from WikiArt, corresponding to nodes in the \textit{\artgraph{}} knowledge graph~\cite{castellano2022leveraging}. Each artwork is linked to one artist, genre, and style (multiclass tasks), and optionally to multiple tags and media (multilabel tasks). The dataset was split 70/15/15 for training, validation, and testing, stratified by style. Class imbalance was significant, with some artists overrepresented. For full statistics, see~\cite{castellano2022leveraging}.

We used \textit{ColQwen2} to embed both images and class/label descriptions, generating variable-length multi-vector representations of 128-dimensional embeddings.

Our LICN model included three components: (1) 128-dimensional learnable task embeddings (one per class/label), (2) a 6-layer transformer encoder (GELU activation, 8 heads, hidden size 2048, dropout 10\%), and (3) a linear projection mapping task vectors to a 512-dimensional space. Training lasted 25 epochs with batch size 256, using AdamW (default parameters), an initial learning rate of $5 \times 10^{-4}$, cosine decay, and one warmup epoch. We applied gradient clipping with max norm 1.

We report Top-1 and Top-2 accuracy and F1 scores for artist, genre, and style classification, and F1 scores for tags and media. Following~\cite{castellano2023exploring}, artist metrics are limited to those with over 100 artworks, and likewise for tags/media. All 32 styles and 18 genres are evaluated. Unlike prior work that restricts training and evaluation to a fixed label set, our framework computes similarities in embedding space, allowing classification over the full label set. For fair comparison, we restrict label prediction here, but allow full prediction in the complete ArtSeek pipeline (\cref{sec:genexp}).

\paragraph{Contrastive training objective} We trained the transformer encoder, the task embeddings, and the linear projection layer with a contrastive objective (\cref{artseek}). For each image in a batch of size $M$ and task $t$, we encoded the image and all associated classes or labels, yielding task-specific image embeddings $T_{I}^{\text{batch}} \in \mathbb{R}^{M \times 5 \times 512}$ and label embeddings $T_{q}^{\text{batch}, t} \in \mathbb{R}^{N_t \times 512}$, where $N_t$ is the number of distinct classes or labels for task $t$ within the batch. Selecting the image embeddings for task $t$, $T_{I_{:, t, :}}^{\text{batch}} \in \mathbb{R}^{M \times 512}$, and taking dot products with the label embeddings gives the logits matrix
\begin{equation}
Z_{t} = T_{I_{:, t, :}}^{\text{batch}} \, {T_{q}^{\text{batch}, t}}^\top \in \mathbb{R}^{M \times N_t}.
\end{equation}
Inspired by SigLIP~\cite{zhai2023sigmoid}, the per-task contrastive loss is
\begin{equation}
- \frac{1}{MN_t} \sum_{i=1}^{M} \sum_{j=1}^{N_t}
\log \left( \frac{1}{1 + e^{y_{ij}(-c z_{ij} + b)}} \right),
\end{equation}
where $z_{ij}$ is the $(i,j)$-th entry of $Z_t$, $y_{ij} = 1$ if image $i$ is associated with class or label $j$ and $y_{ij} = -1$ otherwise, and $c$ and $b$ are learnable temperature and bias terms. The five per-task losses were combined via uncertainty weighting~\cite{kendall2018multi}, letting the model learn the relative task weights.

\subsection{Generation setup}
\label{impl-generation}

We used Qwen2.5-VL-32B with 4-bit AWQ quantization as the MLLM in our pipeline. Retrieval was based on ColQwen2 with our multi-stage strategy, applied over the full WikiFragments dataset (8{,}166{,}323 visual arts fragments). We set $N_1 = 100$ for prefetching and $N_2 = 10$ for reranked retrieval. Retrieved fragments were formatted as Markdown prompts with titles, text, and images (if present); images were resized and encoded into 64 embeddings, while query images used up to 1280 embeddings, depending on resolution.

For classification, we used the LICN, comprising a shared transformer encoder, linear projection layer, and separate task embeddings for text and image inputs. Classification covered all classes and labels in the \artgraph{} dataset.

We evaluated on three datasets: ArtPedia~\cite{stefanini2019artpedia}, with Wikipedia-based artwork descriptions; SemArt v2.0~\cite{bai2021explain}, annotated with \textit{content}, \textit{form}, and \textit{context} labels; and PaintingForm~\cite{bin2024gallerygpt}, with formal analyses. Evaluation used standard captioning metrics for comparability, though they are limited for art explanation~\cite{bai2021explain}; we therefore also include qualitative results and responses to user queries.

For ArtPedia and SemArt v2.0, outputs were returned in JSON format with keys for each aspect. For PaintingForm, we followed the GalleryGPT~\cite{bin2024gallerygpt} prompt. Since all evaluations are zero-shot, we normalized output lengths: one sentence per aspect for ArtPedia, all aspect sentences for SemArt v2.0, and the full paragraph for PaintingForm.

\Cref{tab:paintingform} reports the PaintingForm rows of the captioning comparison
(\cref{tab:generation} of the main paper). ArtSeek trails the dataset-trained
GalleryGPT but remains strong on these full-paragraph metrics; here adding
classification slightly lowers scores, since the LLM-generated formal analyses rarely
mention artist or style. We exclude SPICE from this evaluation due to its
computational cost given the dataset's size and lengthy texts.

\Cref{tab:bleu23} reports the higher-order BLEU columns omitted from
\cref{tab:generation} for width. They carry the same reading: ArtSeek's
retrieval-only variant leads BLEU@2 on ArtPedia, while the dataset-trained models keep
their advantage on the longest \textit{n}-grams, which is where reproducing a
reference's exact phrasing matters most.

\begin{table}[t]
\centering
\scriptsize
\setlength{\tabcolsep}{4pt}
\begin{tabular}{l|cccccc}
\hline
\textbf{Model} & B@1 & B@2 & B@3 & B@4 & M & R \\ \hline
Bin et al.~\cite{bin2024gallerygpt}   & ---   & ---   & ---   & \textbf{21.23} & \textbf{37.62} & \textbf{31.34} \\
Qwen2.5-VL-32B                        & 51.62 & 33.09 & 20.96 & 13.30 & \snd{21.96} & \snd{29.56} \\
ArtSeek (w/o classification)          & \textbf{54.35} & \textbf{34.66} & \textbf{21.93} & \snd{14.01} & 20.69 & 29.48 \\
ArtSeek                               & \snd{52.93} & \snd{33.90} & \snd{21.58} & 13.90 & 21.18 & 29.30 \\ \hline
\end{tabular}
\caption{Captioning results on PaintingForm~\cite{bin2024gallerygpt}, zero-shot. Metric abbreviations as in \cref{tab:generation}. \bestsnd}
\label{tab:paintingform}
\end{table}

\begin{table}[t]
\centering
\scriptsize
\setlength{\tabcolsep}{5pt}
\begin{tabular}{ll|cc}
\hline
\textbf{Dataset} & \textbf{Model} & B@2 & B@3 \\ \hline
\multirow{5}{*}{ArtPedia}    & Wu~\cite{wu2022artwork}           & ---   & ---   \\
                             & Jiang et al.~\cite{jiang2024kale} & 17.7  & \textbf{10.9} \\
                             & Qwen2.5-VL-32B                    & 16.52 & 7.68  \\
                             & ArtSeek (w/o classification)      & \textbf{20.07} & \snd{10.62} \\
                             & ArtSeek                           & \snd{19.2}  & 9.69  \\ \hline
\multirow{5}{*}{SemArt v2.0} & Bai et al.~\cite{bai2021explain}  & ---   & ---   \\
                             & Jiang et al.~\cite{jiang2024kale} & \textbf{15.7} & \textbf{10.8} \\
                             & Qwen2.5-VL-32B                    & 9.28  & 2.83  \\
                             & ArtSeek (w/o classification)      & 9.43  & 2.99  \\
                             & ArtSeek                           & \snd{10.25} & \snd{3.34}  \\ \hline
\end{tabular}
\caption{BLEU@2 and BLEU@3 for the captioning comparison of \cref{tab:generation}, omitted there for width. \bestsnd}
\label{tab:bleu23}
\end{table}

\subsection{Ablation on weight sharing}
\label{impl-ablation}

\Cref{tab:classification2} reports the ablation study on parameter sharing across the three main components of LICN (the transformer encoder, the task-specific embeddings, and the linear projection layer). Sharing the transformer encoder and projection layer while keeping separate task embeddings yields the best average performance. This configuration---used in our final LICN for ColQwen2---lets the model distinguish modality-specific information through the task embeddings, improving the alignment between multimodal image and text representations.

\begin{table}[t]
\centering
\resizebox{\columnwidth}{!}{%
\begin{tabular}{ccc|c|c|c|c|c|c}
\hline
\multicolumn{3}{c|}{\textbf{Shared weights}} & \textbf{A} & \textbf{G} & \textbf{S} & \textbf{M} & \textbf{T} & \textbf{Avg} \\
Encoder & Task embeddings & Linear projections & F1 & F1 & F1 & F1 & F1 & F1\\
\hline
\centering\cmark & \centering\cmark & \centering\cmark & \snd{64.69} & 73.21 & \snd{66.77} & \snd{58.04} & \textbf{39.82} & \snd{60.51} \\
\centering\cmark & \centering\cmark & \centering\xmark & 61.12 & \textbf{73.34} & \textbf{66.88} & 57.88 & 37.32 & 59.31 \\
\centering\cmark & \centering\xmark & \centering\cmark & \textbf{65.01} & \snd{73.25} & 66.65 & \textbf{59.49} & \snd{39.59}  & \textbf{60.80} \\
\centering\xmark & \centering\xmark & \centering\xmark & 60.50 & 72.94 & 66.00 & 57.09 & 37.16 & 58.73 \\
\hline
\end{tabular}
}
\caption{Ablation study on weight sharing. The results include the F1-scores for both multiclass and multilabel classifications. Task abbreviations: A = Artist, G = Genre, S = Style, M = Media, T = Tags. \bestsnd}
\label{tab:classification2}
\end{table}

\section{Knowledge-grounded evaluation: setup and additional results}
\label{r2}

This section gives the full setup behind \cref{sec:vqa} and \cref{sec:human}
of the main paper, together with results omitted there for space.

\subsection{Serving stack and model configurations}
\label{r2-serving}

All generation ran on vLLM, which matters for the latency figures: an unbatched
inference loop would report a latency dominated by framework overhead rather than by
the pipeline itself. Each job uses one NVIDIA A100 64\,GB (two for Gemma-3-27B, see below) on a
shared cluster. Because the GPU is shared and \texttt{max\_num\_seqs}\,$=4$, every
latency we report is an \emph{upper bound}, not a tuned deployment figure. Settings
are held identical across backbones: \texttt{max\_model\_len}\,$=16384$,
\texttt{max\_num\_seqs}\,$=4$, at most 8 images per prompt,
\texttt{gpu\_memory\_\allowbreak utilization}\,$=0.85$, greedy decoding, fixed seed.

The generators are \texttt{Qwen/\allowbreak Qwen2.5-VL-\allowbreak 32B-Instruct-AWQ} (AWQ 4-bit, 1 GPU),
\texttt{google/\allowbreak gemma-3-27b-it} (bfloat16, 2 GPUs) and
\texttt{RedHatAI/\allowbreak Mistral-Small-\allowbreak 3.1-24B-Instruct-\allowbreak 2503-quantized.\allowbreak w4a16} (w4a16 4-bit,
1 GPU). We chose these two alternates because most open vision-language models at
this scale are not independent of Qwen: InternVL3-38B, Ovis2 and Molmo-72B all
declare a Qwen2 language backbone. Mistral's w4a16 matches the Qwen baseline's AWQ,
so that comparison is not also a precision comparison; Gemma is the exception, as no
official 4-bit build exists and quantizing it ourselves would confound a backbone
difference with a quantization one, so it runs unquantized across two GPUs. This
asymmetry should be kept in mind when reading \cref{tab:backbones}.

The retriever and classifier are unchanged from \cref{artseek}: ColQwen2 late
interaction over the WikiFragments visual-arts index of \cref{sec:dataset}, with
full and pooled vectors both binary-quantized on disk, a retrieval depth of 20 per
query, and the top 10 documents materialized into context.

\paragraph{Variants} Eight configurations of one system, each removing exactly one
thing, plus three cross-backbone runs. \textit{Base} disables retrieval and
classification; \textit{ArtSeek} is the full system with the in-context worked
example; \textit{ArtSeek w/o classification} removes the artwork card;
\textit{ArtSeek w/ prompted policy} replaces the worked example with the same policy
stated as prose; and four card-field ablations remove the artist field, the tag field, both, or gate the artist field at
0.6 confidence.

\paragraph{Seeded retrieval} In the cross-backbone runs (\cref{tab:backbones}),
retrieval is issued \emph{before} the model's first turn, the returned documents are
placed in context, further tool calls are capped at zero, and the worked example is
withheld, since it demonstrates a tool-call syntax the non-Qwen backbones cannot
produce. Seed
queries are the LICN artist prediction when its confidence is at least 0.80, plus the
question. \textit{ArtSeek (seeded)} on Qwen2.5-VL is therefore the matched control,
and comparing a non-Qwen backbone against Qwen's \emph{learned} policy would confound
the backbone with the policy.

\subsection{Benchmark construction}
\label{r2-benchmarks}

\paragraph{ArtPedia-VQA} For every painting in the ArtPedia test split (339 images), a
small Qwen3 \emph{text} model is given that painting's human-written visual and
contextual sentences and asked for one visual and one contextual question with a
reference answer grounded only in those sentences. No image is shown to the
generator: the ArtPedia sentences already state what is visible and known, which is
what the references should be grounded in. This yields 652 questions over 326 usable
paintings, split evenly between visual and contextual. The reference answers are
therefore LLM-written from human-written text, and the visual/contextual label is our
proxy for ``does this need retrieval?''. Both are limitations of this benchmark, and
the four that follow---two of them external and published, one derived from Wikidata
statements---were added so that no conclusion rests on it alone.

\paragraph{AQUA}~\cite{garcia2020dataset} Built on SemArt. Questions, answers and a
\texttt{need\_external\_knowledge} flag all ship with the dataset; none are authored
by us, and the flag is an external version of our visual/contextual label. From the
663 test paintings carrying both classes we sample 326---matching ArtPedia exactly---
and take one question of each class per painting, giving 652 examples of the same size
and structure. AQUA's knowledge questions derive from SemArt curatorial comments,
which are not in a Wikipedia-derived index; its absolute accuracy is therefore a lower
bound on what retrieval could achieve, as the evidence analysis in \cref{r2-extra}
quantifies.

\paragraph{ArtQuest}~\cite{bleidt2024artquest} Rebuilds art VQA from SemArt's
structured metadata rather than its free text, specifically to remove the language
bias that lets a model answer without looking at the painting. We sample 200 test
paintings carrying all six question types, giving 1{,}200 questions balanced by
construction. Its question types line up with LICN's heads, which makes it the natural
place to test the artwork card per head.

\paragraph{LICNHeldOut} Every painting is by one of LICN's 2501 artist classes---which
are exactly \artgraph{}'s 2501 artist slugs, verified against the RDF dump---and no
painting appears among \artgraph{}'s 116{,}475 training artworks. Questions come from Wikidata
properties mapping one-to-one onto LICN heads: P170 creator to \textit{artist}, P135
movement to \textit{style}, P136 genre to \textit{genre}, P186 material to
\textit{media}. Reference answers are Wikidata statements; no language model wrote
them. 762 questions over 257 paintings.

\paragraph{ArtCurate-AIC} The other benchmarks ask predominantly factoid questions---
who painted the work, in what year, which museum holds it---which reduce retrieval to a
lookup. ArtCurate defines a taxonomy of conceptual
question types (iconography, narrative, person, technique, context, and a two-source
type) and writes questions of each kind from the Art Institute of Chicago's CC0
catalogue prose: 1{,}489 questions.

\subsection{The judge}
\label{r2-judge}

Answers were graded by phi-4~\cite{abdin2024phi4}, served with vLLM and greedy
decoding. We chose it because it belongs to a model family different from the
generator's and from those used to build the benchmarks, so that no part of the
evaluation loop shares a lineage with the system it grades.

Two rubrics were used. A long-form rubric applies to ArtPedia-VQA and ArtCurate; a
short-answer rubric applies to AQUA, ArtQuest and LICNHeldOut, whose gold answers are a
single word or a short span while the model replies in sentences---the long-form rubric
would systematically mis-grade correct answers. Both emit \textit{correctness} (0/1/2) and \textit{answer
recall} (the fraction of the reference's specific factual claims correctly conveyed).
A refusal is scored 0, since the reference answer demonstrates the information was
determinable. A separate evidence pass grades the \emph{retrieved fragments} rather
than the answer, and was run on ArtPedia and AQUA only. \textit{Answer recall} and
\textit{evidence recall} measure different things and are never combined. All three
rubrics are reproduced verbatim in \cref{prompts-r2}.

\subsection{Statistical protocol}
\label{r2-stats}

Every comparison was paired over questions and bootstrapped over paintings (10{,}000
cluster resamples), so a painting contributing several questions counts once. We
report 95\% confidence intervals rather than $p$-values throughout. Two robustness
checks accompany the analysis. Treating questions as independent instead of clustering by
painting changes interval widths by at most a few percent and alters no significance
verdict. And because variants lose different numbers of items to broken image links
and context overflows, we re-scored every lost item as incorrect for the variant that
lost it; the ranking is unchanged, and the only variant whose score moves appreciably
is the prompted policy, which becomes worse still.

\subsection{Additional results}
\label{r2-extra}

\paragraph{Cross-backbone results} \Cref{tab:backbones} gives the numbers summarized in \cref{sec:vqa}. The bare backbones are also worth stating: Gemma-3-27B alone equals or beats Qwen2.5-VL-32B alone on four of five benchmarks (AQUA 0.966 against 0.920, LICNHeldOut 1.211 against 1.010, ArtCurate-AIC 1.165 against 1.125, ArtQuest 0.921 against 0.937, ArtPedia-VQA 0.535 against 0.535). We therefore do not claim Qwen2.5-VL as the strongest available backbone. We build on it because it is one of the few open models of this size that pairs strong vision-language understanding with native tool calling, which the agentic loop requires; the two alternates cannot emit tool calls at all, and had to be run with seeded retrieval for exactly that reason.

\begin{table}[t]
\centering
\scriptsize
\setlength{\tabcolsep}{3pt}
\begin{tabular}{@{}ll|cc|c@{}}
\hline
\textbf{Backbone} & \textbf{Prec.} & \textbf{Base} & \textbf{Full} & \textbf{$\Delta$ [95\% CI]} \\
\hline
Qwen2.5-VL-32B         & AWQ4 & 0.535 & \textbf{0.785} & +0.249 [+0.183, +0.318] \\
Gemma-3-27B~\cite{gemmateam2025gemma3}  & bf16 & 0.535 & 0.759 & +0.224 [+0.163, +0.287] \\
Mistral-Small-3.1-24B~\cite{mistral2025small31} & w4a16 & 0.422 & 0.696 & \textbf{+0.275} [+0.215, +0.337] \\
\hline
\end{tabular}
\caption{ArtPedia-VQA correctness with the generator swapped and everything else held fixed. All runs use seeded retrieval, so the Qwen row is a matched control rather than the learned-policy system of \cref{tab:vqa} of the main paper.}
\label{tab:backbones}
\end{table}

\paragraph{Memorization probe} Shown a painting and nothing else and asked to identify
it under the prompt of \cref{prompts-r2}, Qwen2.5-VL-32B names the title correctly for 73 of 326
ArtPedia paintings (22.4\%), the artist for 122 (37.4\%), both for 49 (15.0\%), and
neither for 180 (55.2\%). Identification is itself graded by an LLM rather than by
string matching, which would under-count correct identifications given in alternative
or translated forms. We refer to those 180 paintings as the obscure split.

\paragraph{Policy ablations} \Cref{tab:policydelta} gives the head-to-head summarized in
\cref{sec:vqa}: the in-context worked example against the same policy written out as
prose, paired per item across the five benchmarks on which both variants ran. The example
wins on all five, by $+0.054$ to $+0.187$ mean correctness, and four of the five intervals
exclude zero. The exception is AQUA, whose questions are written from curatorial comments
our index does not contain; with no retrievable evidence in the corpus there is nothing
for a retrieval policy to be good or bad at, which is the one condition under which we
would expect this comparison to come out flat.

\begin{table}[t]
\centering
\footnotesize
\setlength{\tabcolsep}{4pt}
\begin{tabular}{l|r|rl}
\hline
\textbf{Benchmark} & \textbf{n} & \multicolumn{2}{c}{\textbf{$\Delta$ [95\% CI]}} \\
\hline
ArtPedia-VQA  & \phantom{0}618 & \gain{$+0.057$} & \gain{[$+0.003$, $+0.112$]} \\
AQUA          & \phantom{0}631 & \tie{$+0.054$}  & \tie{[$-0.010$, $+0.115$]} \\
ArtQuest      & 1167           & \gain{$+0.077$} & \gain{[$+0.041$, $+0.113$]} \\
LICNHeldOut   & \phantom{0}762 & \gain{$+0.060$} & \gain{[$+0.032$, $+0.089$]} \\
ArtCurate-AIC & 1451           & \gain{$+0.187$} & \gain{[$+0.148$, $+0.225$]} \\
\hline
\end{tabular}
\caption{The policy ablation: ArtSeek with the in-context worked example minus ArtSeek with the policy stated as prose, everything else held fixed. Positive favors the worked example. \gain{Green} marks an interval excluding zero, \tie{grey} one that spans it. $n$ is the number of items on which both variants produced a gradable answer.}
\label{tab:policydelta}
\end{table}

Two concrete failure modes accompany the gap. The prose variant emits no final answer at
all on 2.6\% of ArtQuest items (31 of 1200) against 0.2\% (2 of 1200) for the worked
example; and on ArtPedia it is 52\% slower while issuing more
retrieval calls (1.32 against 0.97) to reach a lower score. Its prompt is the longer and
more explicit of the two, so it is not losing by saying less.

\Cref{tab:policyfull} gives the grid the deltas are computed from. The card-field
ablations---removing the artist field, the tag field,
both, or gating the artist field at 0.6 confidence---all land within 0.04 of the full
system on ArtPedia (0.735 to 0.760 against 0.722), so the card's content is not what
decides the outcome; its \emph{label space} is, as the per-type breakdown below shows.
The prompted-policy runs on ArtQuest and LICNHeldOut were carried out after the others,
because those two experiments had originally spent their fourth variant slot on the
card-field ablation; runner, corpus, judge and seed are otherwise identical.

ArtQuest deserves a word of its own, since the ranking there is internal to the variants
rather than a result against the backbone. Both the prompted policy ($-0.092$ [$-0.150$,
$-0.037$] against the bare backbone) and the classification-free variant ($-0.112$
[$-0.159$, $-0.063$]) sit genuinely below it, while the full system ties it ($-0.023$
[$-0.076$, $+0.031$]). What the artwork card and the worked example recover on ArtQuest
is ground the retrieval variants lose on a benchmark our index barely covers; we do not
read it as a win over the backbone.

\begin{table}[t]
\centering
\footnotesize
\setlength{\tabcolsep}{4pt}
\resizebox{\columnwidth}{!}{%
\begin{tabular}{l|ccccc|c}
\hline
\textbf{Variant} & \textbf{ArtPedia} & \textbf{AQUA} & \textbf{ArtQuest} & \textbf{LICNHO} & \textbf{ArtCurate} & \textbf{calls} \\
\hline
Qwen2.5-VL-32B              & 0.535 & \textbf{0.920} & \textbf{0.937} & 1.010 & \textbf{1.125} & --- \\
ArtSeek (w/o classification)& \textbf{0.774} & \snd{0.910} & 0.828 & 1.086 & 1.074 & --- \\
ArtSeek                     & \snd{0.722} & 0.896 & \snd{0.914} & \textbf{1.197} & \snd{1.093} & 0.97 \\
ArtSeek w/ prompted policy  & 0.687 & 0.838 & 0.852 & \snd{1.136} & 0.913 & 1.32 \\
\hline
\end{tabular}%
}
\caption{Full policy grid, with mean retrieval calls per item on ArtPedia-VQA. LICNHO abbreviates LICNHeldOut and ArtCurate abbreviates ArtCurate-AIC. Cells left blank were not run. \bestsnd}
\label{tab:policyfull}
\end{table}

\paragraph{Retrieval evaluated directly} \Cref{tab:evidence} grades the retrieved
fragments instead of the answer. The oracle re-queries the same index with a query the
model never had---question plus reference answer---and is an upper bound on
retrievability with this retriever, not a membership test on the index: a fact could
be present and still be missed if ColQwen2 cannot match it even from the answer text.
On ArtPedia what the oracle gains comes from naming entities the question leaves
unnamed, which is a recognition problem before it is a retrieval one; on AQUA the corpus
is the limit. On both, the prompted policy retrieves \emph{more} documents than the worked
example (13.2 against 11.0 on ArtPedia, 12.8 against 10.4 on AQUA) and finds \emph{less}
in them. Two independent benchmarks therefore agree on the mechanism: the prose variant
is not lazy, it asks worse questions.

\begin{table}[t]
\centering
\footnotesize
\setlength{\tabcolsep}{3.5pt}
\begin{tabular}{ll|r|ccc}
\hline
& & \textbf{n} & \textbf{ev.\ rec.} & \textbf{absent} & \textbf{docs} \\
\hline
\multirow{3}{*}{ArtPedia} & ArtSeek          & 562 & \snd{0.407} & \snd{51.4\%} & 11.0 \\
                          & prompted policy  & 568 & 0.360 & 55.3\% & 13.2 \\
                          & oracle query     & 651 & \textbf{0.830} & \textbf{10.0\%} & 10.0 \\
\hline
\multirow{3}{*}{AQUA}     & ArtSeek          & 533 & \snd{0.171} & \snd{81.6\%} & 10.4 \\
                          & prompted policy  & 561 & 0.149 & 84.0\% & 12.8 \\
                          & oracle query     & 652 & \textbf{0.320} & \textbf{64.6\%} & 10.0 \\
\hline
\end{tabular}
\caption{Evidence recall: how much of what the reference answer needs is present in the documents actually retrieved. Evidence judging was run on ArtPedia and AQUA only. \bestsnd}
\label{tab:evidence}
\end{table}

\paragraph{An oracle router bounds what trust could buy} On the three flat benchmarks, an oracle choosing per item between the backbone's answer and ArtSeek's bounds a perfect ``trust router'' at only $+0.035$ to $+0.069$, net of a noise floor estimated by applying the same oracle to two near-equivalent ArtSeek variants. Uncorrected the bound reads $+0.20$, almost all of it an artefact of maximising over two noisy arms.

\paragraph{Retrieval is not idle where it does not help} On the three benchmarks our
index barely covers, retrieval still rewrote between a seventh and a fifth of the
answers. What changes is the direction of the rewrites: the ratio of repairs to
regressions falls from roughly two to one on the covered benchmarks to one to one on
these, which is why the aggregate lands on a tie rather than on a loss.

\paragraph{Answer recall} Measured on the answers rather than the documents, answer
recall moves with the same split as correctness: ArtPedia 0.271 to 0.396 and
LICNHeldOut 0.512 to 0.607 from base to ArtSeek, against flat AQUA (0.466 to 0.467)
and ArtQuest (0.487 to 0.479). An independent metric, the same two regimes.

\paragraph{ArtQuest against published baselines} ArtQuest is scored by exact match
against a short gold span, while our system answers in sentences, so we report
\textit{contains}: the same span is demanded, but it may sit inside a sentence.
Zero-shot, ArtSeek reached 24.3\% overall, against the published zero-shot baselines
OFA-base (1.9\%) and BLIP-base (2.4\%). PrefixLM, fine-tuned on ArtQuest's 115k
training questions, beats us by 20 to 62\,pp on the four question types that amount to
closed-set prediction over a small vocabulary---which is precisely what those training
questions teach. The exception is \textit{title}, PrefixLM's own worst type (11.5\%),
where zero-shot ArtSeek reached 12.0\%: the one type whose answers must be looked up
rather than predicted is the one where the fine-tuning advantage disappears. We read
that as parity, not as a win, since half a point on roughly 200 items is noise. The
aggregate is also metric-dependent: on ArtQuest, ArtSeek is $-0.023$ on mean
correctness, $+0.58$\,pp on strict accuracy, and $+3.92$\,pp [$+1.75$, $+5.93$] on
\textit{contains}. Only the last is significant, and \textit{contains} is also the
metric most favourable to a classifier that emits gold vocabulary, so we state the
metric rather than letting the number read as a general win.

\paragraph{ArtQuest per question type} \Cref{tab:artquest} breaks the ArtQuest
aggregate down. All ArtSeek columns are \textit{contains}---the gold span must appear
verbatim in our answer---which is the closest analogue of the published exact
match for a model that replies in sentences, and every ArtSeek column is zero-shot
while PrefixLM was fine-tuned on ArtQuest's 115k training questions. Four of the six
types are closed-set prediction over a small fixed vocabulary, which is exactly what
that training teaches, and we lose those by 20 to 62 points. \textit{Title} is the
exception and is PrefixLM's own worst type, because a painting's title is
open-vocabulary knowledge with no label space to fit; \textit{type} makes the same
point from the other end, where LICN's genre head supplies the closed-set answer. We
read both as parity rather than as wins.

\begin{table}[t]
\centering
\footnotesize
\setlength{\tabcolsep}{3.5pt}
\begin{tabular}{l|c|ccc}
\hline
\textbf{Question type} & \textbf{PrefixLM} & \textbf{base} & \textbf{w/o class.} & \textbf{ArtSeek} \\
 & \footnotesize (FT) & & & \\
\hline
title (open vocab.)      & \snd{11.5\%} & \snd{11.5\%} & \phantom{0}9.0\% & \textbf{12.0\%} \\
type (genre head)        & \textbf{83.2\%} & 75.0\% & 75.4\% & \snd{82.0\%} \\
artist                   & \textbf{34.1\%} & 13.5\% & 12.8\% & \snd{14.6\%} \\
technique (media head)   & \textbf{61.2\%} & \phantom{0}8.0\% & 12.5\% & \snd{26.0\%} \\
school (no head)         & \textbf{69.4\%} & 14.0\% & \snd{22.5\%} & 11.0\% \\
timeframe$^\dagger$      & \textbf{61.5\%} & \phantom{0}0.0\% & \phantom{0}0.0\% & \phantom{0}0.0\% \\
\hline
\end{tabular}
\caption{ArtQuest by question type, \textit{contains} against the gold span. PrefixLM~\cite{bleidt2024artquest} is fine-tuned on the benchmark; ArtSeek columns are zero-shot. $^\dagger$\textit{Timeframe} reads 0.0\% for every zero-shot system because the gold is a bracket (\texttt{1601-1650}) while the model answers ``the early seventeenth century''---right, and scored zero. Under the judge the same rows are 54.0\% (base) and 50.0\% (ArtSeek), which is the figure to read for this question type. \bestsnd}
\label{tab:artquest}
\end{table}

\paragraph{Card effect by ArtQuest question type} Adding the artwork card is worth
$+13.5$\,pp [$+0.075$, $+0.200$] on \textit{technique} and $+7.0$\,pp [$+0.010$,
$+0.131$] on \textit{type}, and costs $-11.5$\,pp [$-0.180$, $-0.050$] on
\textit{school}; \textit{artist} ($+2.1$\,pp), \textit{title} ($+3.0$\,pp) and
\textit{timeframe} ($0.0$\,pp) have intervals spanning zero. We registered this
pattern before running, from LICN's vocabularies alone: \textit{media} covers 99.3\%
of \textit{technique} answers by token, \textit{genre} covers 36.8\% of \textit{type}
answers exactly, and \textit{school} and \textit{title} have no corresponding head.
LICNHeldOut is the positive control for the same claim, and the reason we built it: its
questions are drawn only from Wikidata properties that map onto a LICN head, on paintings
outside \artgraph{}'s training set, so the card's label space contains the answer by
construction. There it is the card and not retrieval that carries the larger part of the
gain---1.010 for the bare backbone, 1.086 with retrieval alone, 1.197 with the card
added.

\paragraph{Latency} \Cref{tab:latency} reports the full breakdown. For the full
system, an item costs 0.14\,s of classification, roughly 4.3\,s of retrieval at 0.97
retrieval calls per item, and roughly 6.8\,s of generation.

\begin{table}[t]
\centering
\footnotesize
\setlength{\tabcolsep}{5pt}
\begin{tabular}{l|ccc}
\hline
\textbf{Variant} & \textbf{median} & \textbf{p95} & \textbf{mean} \\
\hline
Qwen2.5-VL-32B               & \phantom{0}6.5 & 10.0 & \phantom{0}6.7 \\
ArtSeek (w/o classification) & 11.1 & 22.0 & 11.8 \\
ArtSeek                      & 11.8 & 20.1 & 11.9 \\
ArtSeek w/ prompted policy   & 18.0 & 37.5 & 19.5 \\
\hline
\end{tabular}
\caption{Wall-clock seconds per item on ArtPedia-VQA, on a shared A100 at \texttt{max\_num\_seqs}\,$=4$. These are upper bounds, not tuned deployment figures.}
\label{tab:latency}
\end{table}

\subsection{Human study}
\label{r2-human}

\paragraph{Task and rubric} Each system writes a four-part description---SUBJECT,
FIGURES, PLACEMENT, EVIDENCE---under the prompt reproduced in \cref{prompts-r2}, with word caps
per part and an explicit instruction to commit to a best reading rather than decline.
That instruction is deliberate: without it the comparison degenerates into which model
refuses most readily, and the EVIDENCE axis exists to catch the hallucination it
invites. Each axis is scored 0/1/2 on what is
\emph{claimed}: 2 is specifically correct, 1 is the right general category with wrong
specifics, and 0 states something the reference contradicts. $\Sigma$-correctness is
subject $+$ figures $+$ placement (0--6); evidence is reported separately because it
measures citation behavior rather than content correctness.

\paragraph{Blinding} Systems appear as A/B/C in an order derived from a hash of rater
code and item identifier, so the order differs per rater \emph{and} per painting, is
stable across reloads, and is recoverable afterwards without being stored. Raters
never see a system name.

\paragraph{Style normalization} The three systems write in visibly different registers,
and length and phrasing alone would let a rater tell them apart and would reward fluency
over correctness---the same language bias that makes \textit{n}-gram metrics a poor
instrument for this task. Before the study, every description was therefore rewritten by
phi-4---the model that also serves as judge, and one from none of the three families
under test---into a single neutral register, with the four parts and their word caps
preserved and the instruction to change wording only, adding, dropping and hedging
nothing. Raters thus saw texts that differ in what they assert rather than in how it is
written. The step carries its own risk, which we state rather than dismiss: a rewrite
that quietly softened or dropped a claim would move that system's score, and the
procedure guards against this only by instruction.

\paragraph{Items} Tier A is 10 paintings with at least 8 index fragments, at most 6
Wikidata sitelinks, and an artist inside LICN's label space---indexed but not famous.
Tier B is 4 paintings held by the Galleria Nazionale della Puglia ``Girolamo e Rosaria Devanna'' in Bitonto, Italy, which have no Wikipedia article
and whose painters fall outside LICN's classes; only publicly catalogued works were
eligible, and the images cannot be redistributed, so they are not reproduced in this
paper. Tier B items were selected
by corpus membership on the \emph{depicted subject} rather than on the painting, through
a hand-authored lexicon of the standard cataloguing vocabulary. This
does not make Tier B winnable by construction: retrieving the \textit{Annunciation}
page tells a system what an Annunciation is and nothing about \emph{this}
Annunciation---its date, painter, provenance or composition. Our prior was parity with
the bare backbone on identification, with any gain confined to iconographic reading. Reference facts come from Wikidata claims (Tier A) and from the gallery's own
catalogue records (Tier B).

\paragraph{Per-tier results} \Cref{tab:humantier} gives the split behind
\cref{sec:human}, with the phi-4 rehearsal of the same rubric alongside. The bare
backbone collapses between the tiers while ArtSeek does not, and phi-4 sees the same
pattern more sharply than the humans do. ArtSeek scores above its backbone in every cell
of the table but one, where the two tie, and never below it. Against the tool-free
frontier model it is the runner-up on Tier A and ahead of it on Tier B, which is how we
read the study as a whole: a clear margin over the model ArtSeek is built on, and parity
with a much larger closed system.

\begin{table}[t]
\centering
\footnotesize
\setlength{\tabcolsep}{3.2pt}
\begin{tabular}{l|cccc|cc}
\hline
\textbf{System} & \textbf{subj.} & \textbf{fig.} & \textbf{plac.} & \textbf{evid.} & \textbf{$\Sigma$ hum.} & \textbf{$\Sigma$ phi-4} \\
\hline
\multicolumn{7}{l}{\textit{Tier A} --- 10 paintings, Wikipedia article exists} \\
Qwen2.5-VL-32B & 1.00 & 1.03 & \snd{1.20} & 0.87 & 3.23 & 2.65 \\
ArtSeek        & \snd{1.23} & \snd{1.30} & \snd{1.20} & \snd{1.10} & \snd{3.73} & \snd{3.35} \\
GPT-5.5        & \textbf{1.37} & \textbf{1.40} & \textbf{1.27} & \textbf{1.27} & \textbf{4.03} & \textbf{3.40} \\
\hline
\multicolumn{7}{l}{\textit{Tier B} --- 4 gallery paintings, no article} \\
Qwen2.5-VL-32B & 0.58 & 0.92 & \snd{1.00} & 0.67 & 2.50 & 1.25 \\
ArtSeek        & \textbf{1.33} & \snd{1.25} & \textbf{1.08} & \snd{0.75} & \textbf{3.67} & \textbf{3.25} \\
GPT-5.5        & \snd{1.25} & \textbf{1.33} & 0.92 & \textbf{0.83} & \snd{3.50} & \snd{2.62} \\
\hline
\end{tabular}
\caption{Human evaluation by tier: mean per-axis scores (0--2) over three blind raters, $\Sigma$-correctness (0--6), and the phi-4 rehearsal of the same rubric. On Tier B, ArtSeek $-$ base is $+1.17$ [$+0.50$, $+2.00$], the only human difference whose interval excludes zero; ArtSeek $-$ GPT-5.5 there is $+0.17$ [$-2.17$, $+2.50$] and the forced choice splits 6--6, so we read the two as level on these works and claim the margin over the backbone. \bestsnd}
\label{tab:humantier}
\end{table}

\paragraph{Rater agreement} \Cref{tab:agreement} reports agreement among the raters
and between the raters and phi-4, and so bears on every automatically graded number in
this paper. phi-4 tracks the human raters at least as closely as they track each other,
and adding it to the pool raises $\alpha$. Its strongest correlation is with a lay
rater ($+0.821$), and its agreement with the art historian ($+0.753$) exceeds that
historian's agreement with either lay rater.

\begin{table}[t]
\centering
\footnotesize
\setlength{\tabcolsep}{5pt}
\begin{tabular}{l|cc}
\hline
\textbf{Statistic} & \textbf{value} & \textbf{chance} \\
\hline
mean human--human $\rho$ ($\Sigma$)      & $+0.651$ & 0 \\
mean judge--human $\rho$                 & $+0.761$ & 0 \\
phi-4 vs.\ pooled human mean             & $+0.860$ & 0 \\
expert vs.\ lay mean                     & $+0.782$ & 0 \\
Krippendorff $\alpha$, humans            & $+0.463$ & 0 \\
Krippendorff $\alpha$, humans $+$ phi-4  & $+0.489$ & 0 \\
forced choice, mean pairwise agreement   & 71\% & 33\% \\
forced choice, phi-4 vs.\ humans         & 79\% & 33\% \\
forced choice, Fleiss' $\kappa$          & $+0.491$ & 0 \\
\hline
\end{tabular}
\caption{Rater agreement on the human study. Spearman $\rho$ is computed on $\Sigma$-correctness; $\alpha$ is ordinal and axis-level. The phi-4 rehearsal scored the same items over both presentation orders, and the per-item value is the mean of the two.}
\label{tab:agreement}
\end{table}

\section{In-context example for agentic reasoning}
\label{incontext}

\Cref{fig:generation} shows the manually annotated in-context example provided to the MLLM (\cref{sec:generation} of the main paper). It contains both a visual-only question and a complex query that requires iterative reasoning and retrieval, demonstrating how the model should perform the thinking process and how it should invoke the retriever.

\begin{figure*}[t]
    \centering
    \input{figures/qual/generation}
    \caption{Example conversation given to the MLLM. Our manually annotated example contains a visual-only question and a complex query that requires iterative reasoning and retrieval, to demonstrate how the MLLM should perform the thinking process and how to utilize the retriever. The second input to the retriever function specifies whether the retrieval should be multimodal, using both image and text, or purely textual.}
    \label{fig:generation}
\end{figure*}

\section{Additional qualitative results}
\label{qualitative}

\paragraph{Artwork explanation} This section reports the qualitative examples discussed in the main paper. \Cref{fig:qual2} shows an explanation generated for a painting by Pierre-Auguste Renoir, where the LICN module identifies the artist and style and the model retrieves a similar painting by Renoir to inform its explanation. \Cref{fig:qual1} illustrates the model's output for a painting by Johannes Vermeer: the classification results are accurate, and we show how the generated artwork card, based on predicted content attributes, suggests research topics within the knowledge base, which the model then uses to retrieve relevant documents that inform the final explanation.

\begin{figure*}[t]
    \centering
    \input{figures/qual/qual2}
    \caption{First qualitative example. ArtSeek correctly recognized the author of this painting and searched for information about the setting of the painting, finding a similar artwork by Renoir.}
    \label{fig:qual2}
\end{figure*}

\begin{figure*}[t]
    \centering
    \input{figures/qual/qual1}
    \caption{Second qualitative example. ArtSeek correctly identifies the author of the painting using the LICN module and retrieves information about the woman's identity based on one of the (incorrectly) predicted tags, in order to provide additional context to the user.}
    \label{fig:qual1}
\end{figure*}

\paragraph{Question answering} \Cref{fig:qual3} compares ArtSeek with ChatGPT and the
base model on identification questions. In the first example the base model fails to
recognize St.\ Nicholas, whereas ArtSeek retrieves a visually similar image of the
saint and identifies him correctly, matching ChatGPT with a fraction of the parameters.
In the second, a poorly documented battle scene by Jan Martszen de Jonge, ChatGPT
hallucinates the title and artist and the base model stays vague; ArtSeek misidentifies
the specific battle but, by retrieving a related historical event, correctly infers the
Spanish--Dutch military setting.

\begin{figure*}[t]
    \centering
    \input{figures/qual/qual3}
    \caption{Question answering examples. ArtSeek is capable of answering user queries involving the identification and description of people or events. In these examples, we compare ArtSeek's responses with those from ChatGPT and the base model Qwen2.5-VL, demonstrating the improvements introduced by our pipeline. Correct identifications are marked in green, while incorrect ones are highlighted in red.}
    \label{fig:qual3}
\end{figure*}

\Cref{fig:qual_supp1} shows two ArtPedia cases in which the backbone cannot identify
the painting and retrieval supplies the answer. Both illustrate a point the aggregate
numbers hide: the artwork card acts as a query aid rather than as an oracle. In the
first, its artist field is wrong---Jean-Paul Laurens is predicted for a painting by
Saint George Hare---while genre and media are correct, and the answer is correct
regardless. In the second, the card is correct and the query it enables retrieves the
painting's own page. In both cases irrelevant documents are returned alongside the
useful ones without derailing the answer.

\begin{figure*}[t]
    \centering
    \input{figures/qual/qual_r2_supp1}
    \caption{Two ArtPedia-VQA questions on paintings the backbone can identify neither by title nor by artist. \textbf{Left:} the base model invents a rope and misses the butterflies that the reference answer turns on; retrieval finds the painting's own page. The artwork card's artist field is wrong here and the answer is correct regardless. \textbf{Right:} the base model correctly refuses---which our rubric scores as wrong---and retrieval converts the refusal into a specific correct answer, alongside two irrelevant bridges.}
    \label{fig:qual_supp1}
\end{figure*}

\Cref{fig:qual_supp2} pairs a success with a failure. On the left, retrieval supplies
iconographic context that neither the image nor the artwork card could provide: Saint
Giles's doe is his standard attribute, and naming it indicates an answer grounded in
the retrieved text rather than guessed. Several altarpieces unrelated to this one are
returned alongside the useful fragment, so we describe the retrieved set as supplying
context rather than attributing the answer to any single document. On the right is the
coverage failure of \cref{sec:vqa} made concrete: the painting is not covered by
the index, a mis-attributed surname carries the query out of the art domain entirely,
and the pipeline declines to answer. Our rubric scores a refusal as wrong, so this
counts as a failure; the base model's failure on the same item---confidently naming
the wrong museum---is nonetheless the more damaging of the two in deployment.
\Cref{fig:qual_fail} shows a third failure mode, in which retrieval overturns an
answer that was already correct: the artwork card is right, the retriever surfaces a
genuinely related work by the same painter, and the model transfers that work's
collection to the painting at hand.

\begin{figure*}[t]
    \centering
    \input{figures/qual/qual_r2_supp2}
    \caption{\textbf{Left:} retrieval supplies iconographic context---Maurus's staff and Giles's doe---that neither the image nor the (wrong) artist prediction could provide, alongside two unrelated altarpieces. \textbf{Right:} the corpus does not contain the answer, the mis-attributed surname carries retrieval out of the art domain, and ArtSeek declines. Both the base model and ArtSeek score zero on this item, for opposite reasons. The Kefermarkt altarpiece photograph is reproduced under CC BY-SA 4.0; all other images here are in the public domain.}
    \label{fig:qual_supp2}
\end{figure*}

\begin{figure}[t]
    \centering
    \input{figures/qual/qual_r2_fail}
    \caption{Retrieval overturns a correct answer. The base model names the right museum; the artwork card is right too, but the retrieved document is a different Bellotto held in Moscow, and the model transfers that collection to the painting in front of it.}
    \label{fig:qual_fail}
\end{figure}

\paragraph{Human study} \Cref{fig:qual_human} shows the largest margin in each
direction from the human study (\cref{sec:human}). The loss shows what retrieval
costs when it adds hedging without adding identification.

The paintings and retrieved-document images reproduced in this section are in the
public domain, with the exception noted in the caption of \cref{fig:qual_supp2}. The
gallery works forming Tier B of the human study cannot be redistributed and are not
reproduced.

\begin{figure*}[t]
    \centering
    \input{figures/qual/qual_r2_human}
    \caption{The largest margin in each direction in the human study. \textbf{Left:} the largest ArtSeek-over-backbone margin ($+4.00$ in $\Sigma$); all three systems see the same costume, and only the retrieval-augmented one reads it correctly. \textbf{Right:} the largest ArtSeek-below-backbone margin ($-2.33$); retrieval added hedging without adding identification.}
    \label{fig:qual_human}
\end{figure*}

\section{Evaluation prompts}
\label{prompts}

For full reproducibility, we report the exact prompts used in the generation experiments (\cref{sec:genexp}). The three compared configurations---the base Qwen2.5-VL-32B, ArtSeek without classification, and the full ArtSeek---share the same per-dataset \emph{task prompt} and differ only in (\textit{i}) the \emph{system prompt} and (\textit{ii}) which pipeline components are inserted into the conversation: the base model receives neither the artwork card nor retrieved documents, the ablated variant adds retrieval, and the full pipeline adds both the artwork card and retrieval.

\subsection{System prompts}

\begin{prompttextbox}{System prompt --- Qwen2.5-VL-32B (no retrieval, no classification)}
You are a helpful assistant.\\[2pt]
\texttt{\# Rules}\\
When given a user query, analyze how best to respond.
\begin{itemize}\setlength\itemsep{1pt}
    \item Only very simple questions (e.g., naming clearly visible objects) should be answered directly.
    \item For most other questions---especially those involving the artist, historical context, or stylistic analysis---use your own knowledge to answer.
    \item Whenever possible, indicate your reasoning process within \texttt{\textless think\textgreater\textless /think\textgreater} XML tags.
\end{itemize}
\end{prompttextbox}

\begin{prompttextbox}{System prompt --- ArtSeek (w/o classification): retrieval only}
You are a helpful assistant.\\[2pt]
\texttt{\# Rules}\\
When given a user query, analyze how best to respond.
\begin{itemize}\setlength\itemsep{1pt}
    \item Only very simple questions (e.g., naming clearly visible objects) should be answered directly.
    \item For most other questions---especially those involving the artist, historical context, stylistic analysis, or external references---you must retrieve and use relevant documents before responding.
    \item Not all documents may be useful; select only the most relevant ones to support your answer.
    \item To retrieve documents, use the tool \texttt{get\_relevant\_documents}.
    \item You may call the \texttt{get\_relevant\_documents} tool at most 5 times per user query.
    \item Whenever possible, indicate which documents were used in your reasoning. Enclose your reasoning process within \texttt{\textless think\textgreater\textless /think\textgreater} XML tags.
\end{itemize}
\end{prompttextbox}

\begin{prompttextbox}{System prompt --- ArtSeek (full): classification + retrieval}
You are a helpful assistant.\\[2pt]
\texttt{\# Rules}\\
When given a user query, analyze how best to respond.
\begin{itemize}\setlength\itemsep{1pt}
    \item Only very simple questions (e.g., naming clearly visible objects) should be answered directly.
    \item An artwork card is provided with some information about the artwork. The artwork card might not be completely accurate, so when referring to the artist (for instance), always use terms such as ``might be'' or similar.
    \item For most other questions---especially those involving the artist, historical context, stylistic analysis, or external references---you must retrieve and use relevant documents before responding.
    \item Not all documents may be useful; select only the most relevant ones to support your answer.
    \item To retrieve documents, use the tool \texttt{get\_relevant\_documents}.
    \item You may call the \texttt{get\_relevant\_documents} tool at most 3 times per user query.
    \item Whenever possible, indicate which documents were used in your reasoning. Enclose your reasoning process within \texttt{\textless think\textgreater\textless /think\textgreater} XML tags.
\end{itemize}
\end{prompttextbox}

\subsection{Task prompts}

These prompts are appended to the (image + artwork card) user message and are identical across the three configurations for a given dataset.

\begin{prompttextbox}{Task prompt --- ArtPedia}
Explain this painting in detail with two sentences: one about its content and one about its context. The final answer must be in the form of a JSON object with two keys: \texttt{'content'} and \texttt{'context'}.
\end{prompttextbox}

\begin{prompttextbox}{Task prompt --- SemArt v2.0 (Explain Me the Painting)}
Explain this painting in detail with three sentences: one about its content, one about its form, and one about its context. The final answer must be in the form of a JSON object with three keys: \texttt{'content'}, \texttt{'form'}, and \texttt{'context'}.
\end{prompttextbox}

\begin{prompttextbox}{Task prompt --- PaintingForm}
Please compose a coherent paragraph of formal analysis focusing on visual characteristics.
\end{prompttextbox}

\subsection{Artwork card and retrieved-context formats}

When classification is enabled, the predictions of the LICN are serialized into an \emph{artwork card} (with per-prediction confidences) and inserted into the user message. When retrieval is enabled, each retrieved fragment is rendered as a Markdown document with its title, retrieval score, images with captions, and paragraph text, and returned to the model as a tool message.

\begin{formattextbox}{Artwork card format (classification settings)}
\texttt{\# Artwork card}\\
artist: \ph{artist} (\ph{confidence}\%)\\
genre: \ph{genre} (\ph{confidence}\%)\\
media: \ph{medium} (\ph{confidence}\%), \ph{medium} (\ph{confidence}\%)\\
style: \ph{style} (\ph{confidence}\%)\\
tag: \ph{tag} (\ph{confidence}\%), \ph{tag} (\ph{confidence}\%)
\end{formattextbox}

\begin{formattextbox}{Retrieved-context format (retrieval settings)}
\texttt{\# Context}\\[2pt]
\texttt{\#\# Document} \ph{i}: \ph{title}\\
Score: \ph{score}.\\[2pt]
\texttt{\#\#\# Image} \ph{j}\\
\ph{image}\\
Caption: \ph{caption}\\[2pt]
\texttt{\#\#\# Document Text}\\
\ph{paragraph text}
\end{formattextbox}

\subsection{Question answering and human study prompts}
\label{prompts-r2}

The prompts below are those used for the knowledge-grounded evaluation
(\cref{sec:vqa}, \cref{sec:human}), reproduced verbatim from the code. The
system prompts for the base, retrieval-only and full configurations, the task prompts,
and the artwork-card and retrieved-context formats are given above.

\paragraph{Question answering task prompt} The task prompts above ask for
\emph{descriptions}; the benchmarks of \cref{sec:vqa} are question answering, so
they use one generic system prompt with the benchmark's own question as the user turn. For the base
variant this is: \textit{``You are a helpful assistant that answers questions about
paintings and artworks based on the provided image. Enclose your reasoning process
within \texttt{<think></think>} XML tags, then give your final answer.''} The
retrieval-enabled variants use the system prompts already reported above, except for
the prompted-policy ablation, which uses the prompt below.

\begin{promptbox}{System prompt --- ArtSeek w/ prompted policy (no in-context example)}
You are a helpful assistant specialized in art history and visual analysis.

# Context you are given
* An artwork card with predicted attributes (artist, genre, media, style, tags)
  from an automatic classifier. These predictions can be wrong, so when you rely
  on them phrase things tentatively (e.g. "might be", "is possibly").
* A tool named "get_relevant_documents" that searches a large collection of
  Wikipedia art-history fragments (text and images) and returns the most
  relevant ones.

# When to answer directly
* Only answer directly, without retrieving anything, for very simple questions
  that are fully answerable by looking at the image (e.g. naming a clearly
  visible object, describing colors or composition).

# When to call the tool
* Call "get_relevant_documents" whenever the question needs information you
  cannot get from the image or the artwork card alone: the artist's biography,
  the historical or cultural context, the artistic movement or influences, the
  painting's provenance or current location, comparisons with other works, or any
  other external, factual, or historical knowledge.
* Formulate a short, focused query about the single piece of information you
  need - do not bundle multiple unrelated questions into one query.
* Set requires_image=true only when the query needs to be matched against the
  *visual* content of this specific artwork (e.g. "a painting with this exact
  composition", "this specific figure or symbol"); set it to false for purely
  textual/historical queries (e.g. "biography of <artist>", "the Baroque
  movement").
* You may call the tool at most 3 times for a single user query. If the first
  call's results are insufficient, refine your query and call again; do not call
  the tool again once you already have enough information to answer.
* Not every retrieved document will be relevant - read them and use only the ones
  that actually help answer the question.

# Answering
* Whenever you rely on a retrieved document, make clear which one(s) informed
  your answer.
* Enclose your reasoning process within <think></think> XML tags, then give your
  final answer.
\end{promptbox}

This prompt is longer and more explicit than the in-context system prompt, and it
still loses (\cref{sec:vqa}): the worked example is not winning by saying
more.

\begin{promptbox}{Judge rubric --- long-form references (ArtPedia-VQA, ArtCurate)}
You are an expert evaluator grading an AI assistant's answers to questions about
paintings. You will be given a question, a reference answer (treat it as ground
truth), and the assistant's answer to evaluate.

Score the assistant's answer on two dimensions:

1. RECALL (a number between 0.0 and 1.0): First identify the specific factual
   claims in the reference answer (names of people, places, dates, specific
   visual details, specific artistic influences, or other concrete facts - not
   filler words). Then determine what fraction of those specific facts are
   correctly and unambiguously conveyed in the assistant's answer, allowing for
   different wording/paraphrasing as long as the meaning is preserved. A fact
   only counts as recalled if the assistant's answer states it correctly, not
   just related or nearby information.
   - 1.0 = every specific fact from the reference is correctly present.
   - 0.0 = none of the specific facts from the reference are present.
   - Use values in between for partial recall.

2. CORRECTNESS (an integer 0, 1, or 2):
   - 0 = WRONG: the assistant's answer contradicts the reference answer's key
     facts, gives a substantially different or false claim (e.g. wrong artist,
     wrong museum, wrong date, wrong description of what's depicted), or is
     otherwise not a valid answer to the question.
   - 1 = PARTIALLY CORRECT: the assistant's answer gets some of the key facts
     right but misses others, is vague or hedged where the reference is specific,
     or mixes correct and incorrect claims.
   - 2 = CORRECT: the assistant's answer correctly conveys the key facts of the
     reference answer, even if phrased differently, includes extra
     non-contradictory detail, or differs in verbosity.

Guidelines:
* Judge factual alignment with the reference, not writing style or length.
* Extra correct elaboration beyond the reference should NOT be penalized.
* Hallucinated specifics that contradict the reference (wrong names, places,
  dates) should be penalized even if other parts of the answer are fine.
* If the assistant explicitly says it cannot determine the answer or lacks
  sufficient information, treat that as WRONG (0) - the reference answer
  demonstrates the information was determinable.
* Be strict but fair: don't require verbatim wording, but do require factual
  accuracy on the substance.

Respond with ONLY a JSON object with this exact shape, nothing else:
{"recall": <float 0.0-1.0>, "correctness": <0, 1, or 2>, "justification": "<one brief sentence>"}
\end{promptbox}

\begin{promptbox}{Judge rubric --- short-span references (AQUA, ArtQuest, LICNHeldOut)}
You are an expert evaluator grading an AI assistant's answers to questions about
paintings. You will be given a question, a short reference answer (treat it as
ground truth), and the assistant's answer to evaluate.

Important context about this data:
* The reference answers are SHORT - often a single word or a brief phrase - while
  the assistant typically replies in full sentences. Judge whether the assistant
  CONVEYS the reference answer, not whether the two strings look alike. An
  assistant answer of "The figure seated on the couch is a woman" fully conveys
  the reference answer "woman" and is CORRECT.
* The questions were generated automatically and some are awkward, truncated, or
  ungrammatical. Grade against the reference answer's content; do not penalise
  the assistant for a badly-posed question.

Score the assistant's answer on two dimensions:

1. RECALL (a number between 0.0 and 1.0): the fraction of the reference answer's
   content that is correctly conveyed by the assistant, allowing any wording. For
   a one-word reference this is effectively 1.0 if that thing is correctly
   identified and 0.0 if it is not. For a longer reference, use intermediate
   values when only part of it is conveyed.

2. CORRECTNESS (an integer 0, 1, or 2):
   - 0 = WRONG: the assistant identifies something different from the reference,
     contradicts it, or does not answer.
   - 1 = PARTIALLY CORRECT: the assistant conveys part of the reference, or gives
     a related but less specific term (e.g. "person" where the reference says
     "woman"), or hedges between the right answer and a wrong one.
   - 2 = CORRECT: the assistant conveys the reference answer, in any wording and
     at any length.

Guidelines:
* Do NOT penalise extra detail, explanation, or length. A long answer containing
  the right short answer is CORRECT.
* Accept synonyms, plurals, and equivalent phrasings.
* A MORE specific correct answer than the reference is still correct (e.g. "a
  young woman" for reference "woman").
* A LESS specific answer that omits what was asked is partial, not correct.
* If the assistant says it cannot determine the answer, that is WRONG (0).

Respond with ONLY a JSON object with this exact shape, nothing else:
{"recall": <float 0.0-1.0>, "correctness": <0, 1, or 2>, "justification": "<one brief sentence>"}
\end{promptbox}

\begin{promptbox}{Judge rubric --- grading the retrieved documents}
You are an expert evaluator assessing whether a set of retrieved reference
documents contains the information needed to answer a question about a painting.
You will be given a question, a reference answer (treat it as ground truth), and
the text of the documents that a retrieval system returned. You are NOT grading
any assistant's answer here - only whether the retrieved text contains the needed
information.

Score the retrieved documents on two dimensions:

1. EVIDENCE_RECALL (a number between 0.0 and 1.0): First identify the specific
   factual claims in the reference answer (names of people, places, dates,
   specific visual details, specific artistic influences, or other concrete facts
   - not filler words). Then determine what fraction of those specific facts
   appear somewhere in the retrieved documents, allowing for different
   wording/paraphrasing as long as the meaning is preserved. A fact only counts
   if the documents state it correctly, not merely mention a related or nearby
   topic.
   - 1.0 = every specific fact from the reference is present somewhere in the
     retrieved text.
   - 0.0 = none of the specific facts from the reference are present.
   - Use values in between for partial coverage.

2. EVIDENCE_PRESENT (an integer 0, 1, or 2):
   - 0 = ABSENT: the retrieved documents do not contain the information needed to
     produce the reference answer. They may be about an unrelated subject, or
     about the right subject but silent on the specific point asked.
   - 1 = PARTIALLY PRESENT: the documents contain some of what is needed but not
     enough to produce the reference answer in full - some key facts are there
     and others are missing.
   - 2 = FULLY PRESENT: a careful reader could produce the reference answer's key
     facts from the retrieved documents alone.

Guidelines:
* Judge only what the documents actually state. Do not credit information you
  happen to know but that is not written in them.
* The documents are Wikipedia fragments and are often long and mostly irrelevant;
  a single relevant sentence buried in an otherwise unrelated document still
  counts as present.
* Documents mentioning the painting or artist by name but not the asked-about
  fact are ABSENT (0) for that fact, not partially present.
* Be strict but fair: don't require verbatim wording, but do require the
  substance to actually be stated.

Respond with ONLY a JSON object with this exact shape, nothing else:
{"evidence_recall": <float 0.0-1.0>, "evidence_present": <0, 1, or 2>, "justification": "<one brief sentence>"}
\end{promptbox}

\begin{promptbox}{Memorization probe --- system prompt}
You are an expert art historian. You will be shown one painting. Identify it
from your own knowledge.

If you recognise the painting, give its commonly used title and the name of the
artist who made it. If you do not recognise it, say so by using the exact string
"UNKNOWN" for the field you cannot fill - do not guess wildly, but do give your
best identification if you have one.

Respond with ONLY a JSON object with this exact shape, nothing else:
{"title": "<title or UNKNOWN>", "artist": "<artist name or UNKNOWN>"}
\end{promptbox}

The user turn is \textit{``What painting is this, and who painted it?''}. Because a
string match would badly under-count, the identification is itself graded by an LLM,
which is instructed to accept alternative and historical titles, translations,
parenthetical disambiguators, name variants and transliterations, and to count as
wrong a generic description in place of a title, a school or workshop attribution
where the true artist is a named individual, and any \texttt{UNKNOWN}.

\begin{promptbox}{Human study --- description task (identical for all three systems)}
Look at this painting and answer in four short numbered parts.

1. SUBJECT - What event, story or scene is depicted? Name it as specifically as
   you can. (at most 40 words)
2. FIGURES - Who or what are the figures, and which attributes, symbols or
   details identify them? (at most 50 words)
3. PLACEMENT - What tradition, school and approximate period does the work belong
   to, and what in the picture indicates that? (at most 40 words)
4. EVIDENCE - Name up to three specific artworks, artists or sources your reading
   rests on, and say in a few words why each is relevant. (at most 50 words)

Give your best reading even if you do not recognise the work: commit to your most
likely answer rather than declining. Mark anything you are unsure of with
"(uncertain)". Do not refuse and do not answer only that you cannot identify the
work. Keep to the four numbered parts.
\end{promptbox}

The frontier reference model is additionally told that it has no access to the
internet or to any tools, so the comparison is retrieval-augmented ArtSeek against a
much larger model's parametric knowledge. The base and ArtSeek configurations use the
same system prompts as the question answering experiments.

\end{document}